\documentclass[fleqn,10pt]{wlscirep}
\usepackage[utf8]{inputenc}
\usepackage[T1]{fontenc}
\usepackage{multirow}
\usepackage{float}
\usepackage{subcaption}
\usepackage{multirow}
\usepackage{todonotes}
\usepackage{nameref}
\usepackage{amsmath,amssymb,amsfonts,bbm}
\usepackage[mathcal]{euscript}
\usepackage[ruled,vlined]{algorithm2e}

\makeatletter
\def\ttl@useclass#1#2{%
  \@ifstar
    {\ttl@labeltrue\@dblarg{#1{#2}}}% {\ttl@labelfalse#1{#2}[]}%
    {\ttl@labeltrue\@dblarg{#1{#2}}}}
\makeatother

\title{Clinical Utility of the Automatic Phenotype Annotation in Unstructured Clinical Notes: ICU Use Cases}

\author[1,5,=]{Jingqing Zhang}
\author[5,=]{Luis Bolanos}
\author[5,=]{Ashwani Tanwar}
\author[2,3]{Julia Ive}
\author[5]{Vibhor Gupta}
\author[1,4,5,*]{Yike Guo}
\affil[1]{Data Science Institute, Imperial College London, London, SW7 2AZ, UK}
\affil[2]{Department of Computing, Imperial College London, London, SW7 2AZ, UK}
\affil[3]{Queen Mary University of London, London, E1 4NS, UK}
\affil[4]{Hong Kong Baptist University, Hong Kong SAR, China}
\affil[5]{Pangaea Data Limited, UK, USA}
\affil[=]{These authors contributed equally to this work.}
\affil[*]{Correspondence to: y.guo@imperial.ac.uk}

\begin{abstract}

\paragraph{Objective}

Clinical notes contain information not present elsewhere, including drug response and symptoms, all of which are highly important when predicting key outcomes in acute care patients. We propose the automatic annotation of phenotypes from clinical notes as a method to capture essential information, which is complementary to typically used vital signs and laboratory test results, to predict outcomes in the Intensive Care Unit (ICU).

\paragraph{Methods}

We develop a novel phenotype annotation model to annotate phenotypic features of patients which are then used as input features of predictive models to predict ICU patient outcomes. We demonstrate and validate our approach conducting experiments on three ICU prediction tasks including in-hospital mortality, physiological decompensation and length of stay for over 24,000 patients by using MIMIC-III dataset.

\paragraph{Results}

The predictive models incorporating phenotypic information achieve 0.845 (AUC-ROC) to predict in-hospital mortality, 0.839 (AUC-ROC) for physiological decompensation and 0.430 (Kappa) for length of stay, all of which consistently outperform the baseline models leveraging only vital signs and laboratory test results. Moreover, we conduct a thorough interpretability study, showing that phenotypes provide valuable insights at the patient and cohort levels. 

\paragraph{Conclusion}

The proposed approach demonstrates phenotypic information complements traditionally used vital signs and laboratory test results, improving significantly forecast of outcomes in the ICU.

\end{abstract}

\begin{document}

\flushbottom
\maketitle

\thispagestyle{empty}

%\noindent Please note: Abbreviations should be introduced at the first mention in the main text – no abbreviations lists. Suggested structure of main text (not enforced) is provided below.

\section*{Summary}

What is already known?

\begin{itemize}

\item Previous works have demonstrated good performance for prediction of outcomes in the Intensive Care Unit (ICU) using bedside measurements and laboratory test results

\item Contextual embeddings from recent Transformer-based Natural Language Processing models have enabled a more accurate detection of medical concepts 

\end{itemize}

\noindent What does this paper add?

\begin{itemize}

\item This paper introduces a new methodology to incorporate contextualized phenotypic features from clinical text and their persistency into the modelling of ICU time-series prediction tasks

\item We conduct an interpretability study, illustrating why and how phenotypic features are highly relevant for ICU outcomes prediction

\end{itemize}

\section*{Introduction}

The accumulation of healthcare data today has reached unprecedented levels: NHS datasets alone record billions of patient interactions every year~\cite{nhs-report}. In particular, due to the close monitoring of patients in an Intensive Care Unit (ICU), a wealth of data is generated for each patient\cite{Johnson2016}, with some information being recorded every minute. 
% Finding efficient ways to use the data from s (EHRs) is crucial to reduce the burden in hospitals and improve patient outcomes. 

In the typical setting, an Electronic Health Record (EHR) contains two types of information, which are structured (e.g. blood tests, temperature, lab results) and unstructured information (e.g. nursing notes, radiology reports, discharge summaries), with the latter composing the biggest part (typically, up to 80\%~\cite{Kong2019}). Both types of information are valuable for the ICU monitoring. The majority of recent research \cite{harutyunyan2019multitask, Subudhi2021, Alves2019} relies though on more straightforward structured information, typically being laboratory test results and vital signs. 

% On the other hand, unstructured information is underexplored and typically two approaches are applied: raw clinical text is input into a classifier for end-to-end prediction on its basis ~\cite{Krishnan2018, Grnarova2016, Marafino2018, Yu2020} or the text is used to extract categorical features at the first step which are then used for prediction. Those categorical features could be pre-extracted medical concepts: e.g., diseases\cite{Cooley2021}.

Among the unstructured data, the phenotype\footnote{In the medical text, the word ``phenotype'' refers to deviations from normal morphology, physiology, or behaviour \cite{robinson2012deep}.} has been received the least attention for the ICU monitoring~\cite{Cooley2021}. This is mainly due to the challenge to extract the phenotypic information expressed by a variety of contextual synonyms. For example, such a phenotype as \textit{Hypotension} can be expressed in text as ``drop in blood pressure'' and ``BP of 79/48''. However, the phenotypes are crucial for understanding disease diagnosis, identifying important disease-specific information, stratifying patients and identifying novel disease subtypes \cite{Liu2019}. 

% There exists a range of clinical Natural Language Processing (NLP) approaches that allow to extract phenotypic information from clinical text in an unsupervised manner~\cite{arbabi2019ncr}. There is thus a possibility to exploit this information in an automatic way, for example, to predict mortality in septic patients \cite{Cooley2021}.

Our work thoroughly investigates the value of phenotypic information as extracted from text for ICU monitoring. We automatically extract mentions of phenotypes from clinical text using a self-supervised methodology with recent advancements in clinical NLP - contextualized word embeddings \cite{alsentzer-etal-2019-publicly} that are particularly helpful for the detection of contextual synonyms. We extract those mentions for over 15,000 phenotype concepts of the Human Phenotype Ontology (HPO)~\cite{Kohler2021hpo}. We enrich the phenotypic features extracted in this manner with the information coming from the structured data (i.e., bedside measurements and laboratory test results). To provide interpretation into our results we use SHAP values\cite{Lundberg2017original}.
% , a methodology that comes from the game theory and that ranks the inputs of a system according to their importance for each individual prediction. 
% With the help of this methodology, we are able to provide highly relevant explanations for a complete picture of a patient's status both at the individual and cohort levels.

We benchmark our approach on the following three mainstream ICU tasks following the practice \cite{harutyunyan2019multitask} for comparison: length of stay, in-hospital mortality and physiological decompensation. 

% Length of stay is one of the major drivers for operative costs and resource allocation in hospitals. Understanding it in advance can be highly useful for decision-making and better resource planning. The task of in-hospital mortality prediction is important for early identification of patients at-risk and has been a focus of the research interest for quite a while, as evidenced by the existence of multiple scoring systems designed to tackle the task (e.g., APACHE\cite{Knaus1991apache}, SAPS\cite{LeGall1993saps}). Knowing which patients are more vulnerable and prone to decompensation (physiological decompensation task) allows clinicians to better monitor and track these patients. Many warning scores methods have been developed for this task as well (e.g., NEWS\cite{Smith2019news}). 

Our {\bf main contributions} are: 
(i) approach to incorporate phenotypic features into the modelling of ICU time-series prediction tasks; 
(ii) investigation of the importance of the phenotypic features in combination with structured information for the prediction of patient course at micro (individual patient) and macro (cohort) levels;
(iii) thorough interpretability study demonstrating the importance of phenotypic features and structured features for the ICU cases;
(iv) demonstration of the utility of automatic phenotyping for ICU use cases.

\section*{Methodology}
\label{sec:method}

\subsection{Data preprocessing}

In this study, we use the publicly available ICU database MIMIC-III~\cite{Johnson2016mimic} and follow the common practice \cite{harutyunyan2019multitask} to define the three ICU tasks, data collection and data preprocessing. We formulate the \textit{in-hospital mortality} problem as a binary classification at 48 hours after admission, in which the label indicates whether the patient dies before discharge. We formulate the problem of \textit{physiological decompensation} as a binary classification, in which the target label corresponds to whether the patient will die in the next 24 hours. We cast the \textit{length of stay} (LOS) prediction task as a multi-class classification problem, where the labels correspond to the remaining length of stay. Possible values are divided into 10 bins, one for the stays of less than a day, 7 bins for each day of the first week, another bin for the stays of more than a week but less than two, and the final bin for stays of more than two weeks. 

For data collection, we use both structured data (e.g. bedside measurements) and unstructured data (e.g. clinical notes) following the filtering criteria \cite{harutyunyan2019multitask} for the patients, admissions and ICU stays in all three tasks. In addition, we discard all the ICU episodes in which a clinical note is not recorded. This reduces our train and test data as compared to the benchmark \cite{harutyunyan2019multitask}, so we recalculate the baseline scores using their code on our new test set for fair comparison. Overall, there are over 24,000 patients in total and the exact numbers of patients, ICU episodes and timesteps per task are reported in Table \ref{table:data_distribution}. Mortality rate across all patients is 13.12\% and decompensation rate across all timesteps is 2.01\%. Most patients stay in ICU less than 7 days, and the distribution of ICU stays per LOS class is presented in detail in Table \ref{table:los_classes}.

For data preprocessing of structured data, we follow the steps\footnote{Accessed in November 2021: \href{https://github.com/YerevaNN/mimic3-benchmarks}{https://github.com/YerevaNN/mimic3-benchmarks}} to collect 17 clinical features (i.e., capillary refill rate, diastolic blood pressure, fraction inspired oxygen, Glasgow coma scale eye opening, Glasgow coma scale motor response, Glasgow coma scale verbal response, Glasgow coma scale total, glucose, heart rate, height, mean blood pressure, oxygen saturation, respiratory rate, systolic blood pressure, temperature, weight, and pH). For data preprocessing of unstructured data, we collect all clinical notes including nursing notes, physician notes and discharge summaries at all timesteps during ICU stays and we observe there is high data sparsity as clinical notes are recorded roughly every 12 hours. The processed structured data and unstructured data are then used as inputs to our approach.

% These phenotypes are generally chronic in nature which reflects a patient's medical background such as cancer and tuberculosis. These conditions are unlikely to change throughout the patient's ICU stay. In practice, this means that a persistent phenotype is made available throughout the entire stay starting from the point it firstly appears. All temporary phenotypes and those that generally last long, but which might resolve in some cases, are marked `transient' to prevent a resolved phenotype from being carried forward too far. The latter phenotypes are temporary and can be acquired or improved during an ICU stay such as pain, fever, cough, etc. For a transient phenotype, we make it present from the moment it appears until a new clinical note appears. 

\subsection*{Algorithm development and analysis}

The proposed approach consists of two steps. The first step is to collect clinical features (more specifically, phenotypic features, standardised by Human Phenotype Ontology (HPO) \cite{Kohler2021hpo}) from unstructured data by using Natural Language Processing (NLP) algorithms. The second step is to combine the phenotypic features from unstructured data and the 17 clinical features from structured data as input features for machine learning classifiers to predict in-hospital mortality, physiological decompensation and LOS in separate.

First, to extract phenotypes from free-text clinical notes, \textbf{we develop a state-of-the-art phenotyping model, which leverages contextualized word embeddings and data augmentation techniques} (paraphrasing and synthetic text generation) to capture names, synonyms, abbreviations and, more importantly, contextual synonyms of phenotypes. For example, ``drop in blood pressure'' and ``BP of 79/48'' are both contextual synonyms of \textit{Hypotension (HP:0002615)}. As a result of the contextual detection of phenotype, the phenotyping model demonstrates superior performance than alternative phenotyping algorithms. We refer the readers to the work \cite{zhang-etal-2021-self} for methodological details. For comparison, we also use alternative phenotyping methods including ClinicalBERT \cite{alsentzer-etal-2019-publicly} (fine-tuned for phenotyping) and NCR~\cite{arbabi2019ncr}. NCR uses a convolutional neural network (CNN) to assign similarity scores to HPO concepts of phrases encoded by using pre-trained non-contextualized word embeddings. 

% We also replace all the phenotypes with their immediate parent using the Human Phenotype Ontology (HPO) \cite{Kohler2021hpo} hierarchy (unless the parent is ``Phenotypic abnormality'' which is the root node of phenotypes in HPO) to reduce the overall feature dimensionality. Overall, this produces a reduction of 40-60\% in the number of features for each patient.

% We divide the phenotypes into persistent and transient by the annotations of an expert clinician. If a phenotype is clinically deemed likely to last an entire admission in the vast majority of typical cases (e.g., tuberculosis, cancer), it is marked as `persistent'. In contrast, if the phenotype can be acquired or improved during an ICU stay, such as pain, fever, cough, it is marked as `transient'. We make transient and persistent phenotypes present from the moment it appears until a new clinical note appears, and until the end of the ICU stay, respectively.

Second, the phenotypic features are combined with structured clinical features together as input features to machine learning classifiers for prediction of the three ICU tasks. We use standard machine learning classifiers: Random Forest (RF)~\cite{Breiman2001}, and Long Short-Term Memory Network (LSTM)~\cite{Hochreiter1997lstm} for prediction. \textbf{We distinguish the phenotypic features between persistent and transient ones to reduce feature sparsity.} More precisely, if a phenotype is clinically deemed likely to last an entire admission in the vast majority of typical cases (e.g., tuberculosis, cancer), it is marked as `persistent'. In contrast, if the phenotype can be acquired or improved during an ICU stay, such as pain, fever, cough, it is marked as `transient'. We make transient and persistent phenotypes present from the moment it appears until a new clinical note appears, and until the end of the ICU stay, respectively. We find this beneficial and will discuss it in Section \ref{subsec:phenotype_persistency}. We also address data sparsity by aggregating HPO terms into their parents (according to the HPO hierarchy).

\subsection*{Evaluation Metrics}

To compare with the previous study \cite{harutyunyan2019multitask}, we use Area Under the Curve of Receiver Operating Characteristic (AUC-ROC) \cite{Lasko2005aucroc} and Area Under the Curve of Precision-Recall (AUC-PR) for In-Hospital Mortality and Physiological Decompensation tasks. We primarily rely on AUC-ROC for statistical analysis as it is threshold independent and used by the benchmark~\cite{harutyunyan2019multitask} as the primary metric. For the LOS task, we use Cohen's Kappa~\cite{Cohen1960kappa} and Mean Absolute Deviation~\cite{PhamGia2001mad} (MAD) with primarily relying on the Kappa scores for statistical analysis.

\subsection*{Model evaluation and statistical analysis}

We use a train-test split based on the benchmark\cite{harutyunyan2019multitask}, but exclude patients without clinical notes, resulting in 21,346 and 3,824 patients for train and test set, respectively. Further, we perform 4-fold cross validation on the training set. All splits are deterministic, so that all the classifiers with different data settings are trained and evaluated with the same subsets of data. We use the bootstrap resampling following the benchmark for statistical analysis of the scores. To compute confidence intervals on the test set we resample it 1,000 times for length of stay and decompensation, and 10,000 times for in-hospital mortality task. Then, we compute the scores on the resampled data to calculate 95\% confidence intervals. 

To provide interpretability and insights into model predictions, we use SHAP values\cite{Lundberg2017original}, the implementation details of which are explained more in Appendix \ref{subsec:shap}. The SHAP values are typically used to explain black box models, and allow us to quantify the importance of a feature and whether it impacts positively or negatively the outcome.

% \subsection*{Data availability}

% The MIMIC-III database is available on PhysioNet repository \cite{physionet2020mimiciii}. The code to generate the three ICU benchmarks from MIMIC-III is published by the study \cite{harutyunyan2019multitask}.

\section*{Results}

\subsection*{Phenotyping}

Across all three tasks, ClinicalBERT finds 664 phenotypes, NCR finds 1,441 phenotypes, and our methodology finds 1,446, in average. 30\% of these phenotypes are persistent (on average across tasks).

\subsection*{Quantitative results}

In general, we investigate the performance of two classifiers: Random Forest (RF) and LSTM. For each of them we investigate the following set of features: structured features only (S) and structured features enriched with phenotypic features coming from one of the three phenotype annotators (ClinicalBERT, NCR, ours).

The main results are presented in Table \ref{table:results_core} and the results from statistical tests are presented in Table \ref{table:results_stat_test}. Overall, they show that phenotypic information complements positively the structured information to improve performance on all tasks.
% (on average 2.16\% for in-hospital mortality, 0.53\% for decompensation, and 0.83\% for LOS across annotators and classifiers).
The improvements with our phenotyping model are statistically significant across all tasks compared against using structured features only or alternative phenotyping algorithms, except for In-Hospital Mortality with RF.
% Improvements from NCR are not significant for physiological decompensation with RF, and LOS with RF. ClinicalBERT did not produce significant improvements for in-hospital mortality with RF and LSTM, physiological decompensation with RF, and LOS with RF. 

This is explained by the fact that phenotypes carry highly valuable information, including response to therapy, development of complication, comorbidities and unmeasured indicators of illness severity, all of which are fundamental to correctly estimate the LOS and mortality risk of a patient~\cite{Kramer2017los, Forte2019}.

% For the particular case of LOS prediction, this is explained by the fact that phenotypic information brings into the picture highly valuable information such as response to therapy, development of complications and unmeasured indicators of severity illness, all of which are not captured by structured information nor by severity scores based on day 1 forecasts, and are fundamental to correctly estimate LOS~\cite{Kramer2017los}. For in-hospital mortality, improvements are explained by the fact that phenotypic information brings into the picture the patient's comorbidities, which are known to be essential to accurately predict the mortality risk~\cite{Forte2019}.

% LSTMs~\cite{Hochreiter1997lstm} consistently outperform other algorithms due to their intrinsic ability to capture temporal relations, presenting average increases of 1.41\%, 0.52\% and 1.51\% for in-hospital mortality, physiological decompensation and LOS, respectively. Setups across all classifiers with phenotypic features as extracted by our phenotyping model have tendency to outperform other similar setups on in-hospital mortality by 0.66\%, on physiological decompensation by 0.42\% and on LOS by 2.67\%.

% Note that clinically validated scores such as APACHE-III and SAPS-II are not competitive across settings.

% Across all three tasks there is no clear winner between NCR and ClinicalBERT. Each of them can outperform the other without any clear tendency. Also, we find that the mean scores from cross-validation aggregate are comparable to the respective test set scores across all the tasks for all the models.  

\begin{table}[!htbp]
\begin{subtable}[c]{1\textwidth}
\centering
\begin{tabular}{|c|c|c|c|}
\hline
\multicolumn{1}{|l|}{Classification   Model} & \multicolumn{1}{l|}{Features Design} & AUC-ROC $\uparrow$            & AUC-PR   $\uparrow$           \\ \hline
SAPS-II                                      & -                                    & 0.756                         & 0.312                         \\ \hline
APACHE-III                                   & -                                    & 0.733                         & 0.308                         \\ \hline
\multirow{4}{*}{Random   Forest}             & S                                    & 0.800 (0.775, 0.824)          & 0.339 (0.286, 0.395)          \\ \cline{2-4} 
                                             & S + NCR                              & 0.828 (0.802, 0.853)          & \textbf{0.467 (0.404, 0.529)} \\ \cline{2-4} 
                                             & S + CB                               & 0.812 (0.787, 0.838)          & 0.403 (0.345, 0.463)          \\ \cline{2-4} 
                                             & S + Ours                             & \textbf{0.845 (0.826, 0.873)} & 0.462 (0.404, 0.524)          \\ \hline
\multirow{5}{*}{LSTM}                        & S \cite{harutyunyan2019multitask}    & 0.825                         & 0.410                         \\ \cline{2-4} 
                                             & S                                    & 0.826 (0.801, 0.848)          & 0.391 (0.334, 0.452)          \\ \cline{2-4} 
                                             & S + NCR                              & 0.841 (0.818, 0.864)          & 0.453 (0.393, 0.513)          \\ \cline{2-4} 
                                             & S + CB                               & 0.826 (0.802, 0.849)          & 0.414 (0.355, 0.476)          \\ \cline{2-4} 
                                             & S + Ours                             & \textbf{0.845 (0.823, 0.868)} & \textbf{0.464 (0.405, 0.523)} \\ \hline
\end{tabular}
\caption{In-hospital mortality}
\label{table:mortality}
\end{subtable}

\begin{subtable}[c]{1\textwidth}
\centering
\begin{tabular}{|c|c|c|c|}
\hline
\multicolumn{1}{|l|}{Classification   Model} & \multicolumn{1}{l|}{Features Design} & AUC-ROC $\uparrow$            & AUC-PR   $\uparrow$           \\ \hline
\multirow{4}{*}{Random   Forest}             & S                                    & 0.826 (0.821, 0.831)          & 0.130 (0.123, 0.138)          \\ \cline{2-4} 
                                             & S + NCR                              & 0.825 (0.821, 0.830)          & 0.124 (0.118, 0.131)          \\ \cline{2-4} 
                                             & S + CB                               & 0.826 (0.821, 0.830)          & 0.125 (0.118, 0.132)          \\ \cline{2-4} 
                                             & S + Ours                             & \textbf{0.845 (0.840, 0.850)} & \textbf{0.180 (0.171, 0.190)} \\ \hline
\multirow{5}{*}{LSTM}                        & S \cite{harutyunyan2019multitask}    & 0.809                         & 0.125                         \\ \cline{2-4} 
                                             & S                                    & 0.824 (0.819, 0.829)          & 0.126 (0.119, 0.133)          \\ \cline{2-4} 
                                             & S + NCR                              & 0.834 (0.829, 0.839)          & 0.134 (0.127, 0.142)          \\ \cline{2-4} 
                                             & S + CB                               & 0.833 (0.828, 0.838)          & 0.114 (0.108, 0.119)          \\ \cline{2-4} 
                                             & S + Ours                             & \textbf{0.839 (0.834, 0.844)} & \textbf{0.145 (0.138, 0.153)} \\ \hline
\end{tabular}
\caption{Physiological decompensation}
\label{table:decompensation}
\end{subtable}

\begin{subtable}[c]{1\textwidth}
\centering
\begin{tabular}{|c|c|c|c|}
\hline
\multicolumn{1}{|l|}{Classification   Model} & \multicolumn{1}{l|}{Features Design} & Kappa  $\uparrow$             & MAD $\downarrow$              \\ \hline
\multirow{4}{*}{Random   Forest}             & S                                    & 0.390 (0.388, 0.392)          & 136.8 (136.2, 137.4)          \\ \cline{2-4} 
                                             & S + NCR                              & 0.390 (0.388, 0.392)          & 142.5 (141.9, 143.1)          \\ \cline{2-4} 
                                             & S + CB                               & 0.376 (0.374, 0.379)          & 144.3 (143.7, 144.9)          \\ \cline{2-4} 
                                             & S + Ours                             & \textbf{0.420 (0.418, 0.422)} & \textbf{110.3 (109.3, 111.3)} \\ \hline
\multirow{5}{*}{LSTM}                        & S \cite{harutyunyan2019multitask}    & 0.395                         & 126.7                         \\ \cline{2-4} 
                                             & S                                    & 0.380 (0.377, 0.382)          & 157.0 (156.3, 157.6)          \\ \cline{2-4} 
                                             & S + NCR                              & 0.406 (0.404, 0.408)          & 123.3 (122.8, 123.9)          \\ \cline{2-4} 
                                             & S + CB                               & 0.388 (0.386, 0.390)          & 120.1 (119.6, 120.6)          \\ \cline{2-4} 
                                             & S + Ours                             & \textbf{0.430 (0.427, 0.432)} & \textbf{116.7 (116.1, 117.2)} \\ \hline
\end{tabular}
\caption{Length of Stay}
\label{table:los}
\end{subtable}

\caption{Results for (a) In-Hospital Mortality, (b) Physiological Decompensation, and (c) Length of Stay. Test set scores are shown with 95\% confidence intervals in brackets if applicable. The best score for each classifier is highlighted in bold. The first row of LSTM refers to scores reported in previous literature, while the second row regards scores reproduce in this study with a comparable cohort. Here, S refers to Structured, NCR to Neural Concept Recognizer\cite{arbabi2019ncr}, CB to ClinicalBERT, and Ours to our phenotyping model.}
\label{table:results_core}
\end{table}

\section*{Discussion}

% In this paper we work on three popular tasks in the ICU settings. Our work shows that these tasks are benefited by the inclusion of phenotypic information found inside clinical notes irrespective of the classification algorithms. Further, we show that our algorithm outperforms state-of-the-art models.

While we conclude that phenotypic information provides useful information to correctly conduct the three ICU tasks, decision support systems in the healthcare domain should be reliable, interpretable and robust. Therefore, we accompany the above results with a thorough study on interpretability, providing explanations both at the patient and cohort levels for the observed predictions, and an assessment of robustness by studying performances across disease-specific sub-cohorts.

\subsection*{Phenotype persistency}
\label{subsec:phenotype_persistency}

We find it beneficial to propagate phenotypes forwards in time. More precisely, each phenotypes is marked by one human clinical expert based on whether it would typically persist throughout an entire ICU stay or not. Consequently, transient (e.g., fever, cough, dyspnea) and persistent (e.g., diabetes, cancer) phenotypes are propagated until the appearance of a new clinical note or the end of the ICU stay, respectively. We perform an ablation study and observe the phenotype propagation is more beneficial to Random Forest (RF) than LSTM. The RF models with phenotype propagation achieve 4.6\% higher AUC-ROC for in-hospital mortality, 2.5\% higher AUC-ROC for decompensation and 3.4\% higher Kappa for LOS than RF without phenotype propagation. However, the LSTM models with phenotype propagation achieve 1.4\% higher AUC-ROC for in-hospital mortality, comparable results for decompensation and 1.1\% lower Kappa for LOS. We hypothesise this is because LSTM by design can better capture temporal relationship given a large amount of data to learn from. The full results can be found in Table \ref{table:ablation}. We believe further investigation focused on learning persistency of phenotypes would be beneficial, not only to boost prediction accuracy, but also to provide insights about temporal duration of phenotypes in the ICU.

\subsection*{Phenotype importance}

\begin{figure}[!htbp]
     \centering
     \begin{subfigure}[b]{0.8\textwidth}
         \centering
         \includegraphics[width=\textwidth]{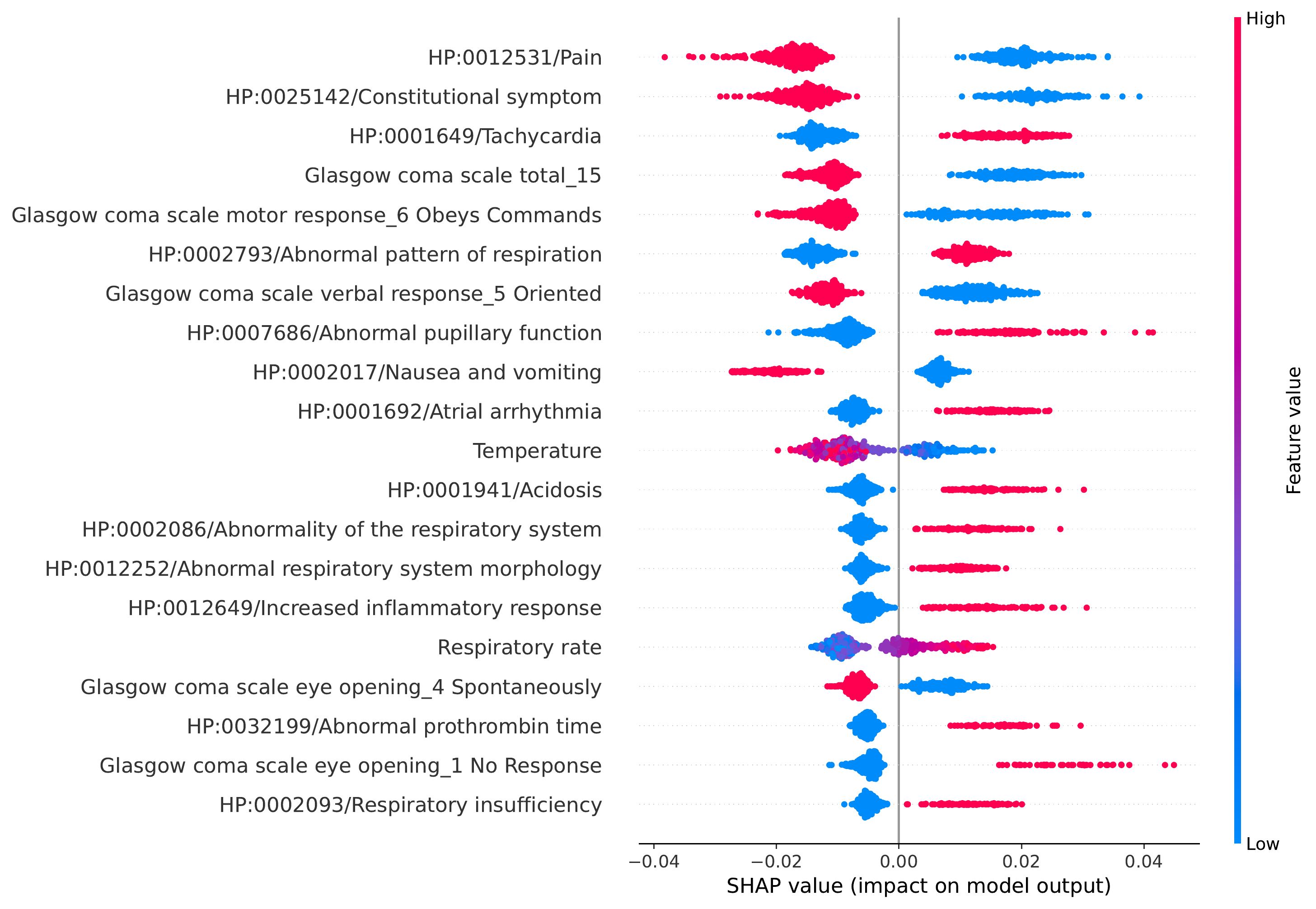}
         \caption{In-Hospital Mortality}
         \label{fig:top_features_mortality}
     \end{subfigure}
     \hfill
     \begin{subfigure}[b]{0.8\textwidth}
         \centering
         \includegraphics[width=\textwidth]{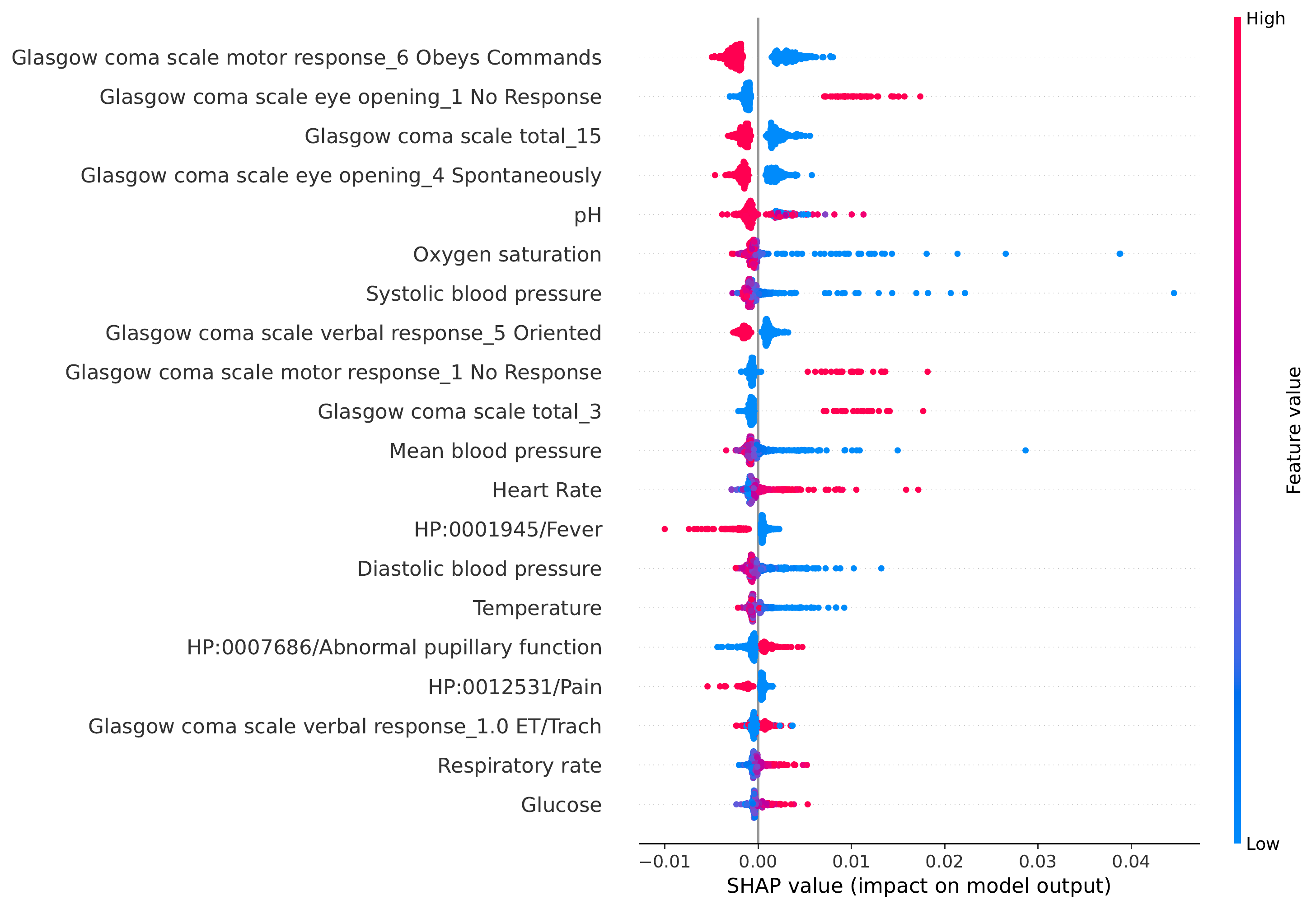}
         \caption{Physiological Decompensation}
         \label{fig:top_features_decomp}
     \end{subfigure}
    \caption{Top features for in-hospital mortality and physiological decompensation . Features are sorted in decreasing importance according to their mean absolute SHAP values. Each row presents a condensed summary of the feature's impact on the prediction. Each data sample is represented as a single dot in each row, and its colour on a particular row represents the value of that sample for that feature, with blue corresponding to lower values or absence, and red to higher ones or presence. The SHAP value (horizontal position of a dot) measures the contribution of that feature on a sample, towards the prediction (right corresponding to mortality or decompensation, and left to survival or out of decompensation risk). For instance, in (a) top row, since the vertical axes clearly splits patients by colour, manifesting \textit{HP:0012531 Pain} consistently leads to lower chances of dying.}
    \label{fig:top_features_beeswarm}
\end{figure}

To further understand the contribution of phenotypic features to the prediction performance, we have studied the most important features with the help of SHAP values\cite{Lundberg2017original}. This analysis and all involving SHAP values are conducted on the Random Forest (RF) models. An illustration of our investigation is in Figure \ref{fig:top_features_beeswarm}, where we present the top predicting features for in-hospital mortality and physiological decompensation. It confirms that phenotypic features are particularly helpful for the in-hospital mortality prediction, given that 13 out of the 20 most important features are phenotypes. This is explained by the fact that 
% since the prediction should comprehend the entire stay rather than predict whether the patient will die within the next 12 hours, 
forecasts need to rely on information that is able to provide insights accurately into the long-term future. 

Contrary to bedside measurements which may not correlate well with future outcomes due to their dynamic nature, phenotypes are highly informative given that they capture, for instance, comorbidities, which are essential for predicting mortality~\cite{Forte2019}. Furthermore, another study \cite{Nielsen2019} including 230,000 ICU patients found that combining the comorbidities with acute physiological measurements yielded the best results, outperforming all mortality scores (APACHE-II, SAPS-II).

% The top ranking feature for mortality prediction is, unexpectedly, whether the patient experiences pain or not. Although not decisive, there is some initial evidence corroborating the fact that pain management improves outcomes in the ICU~\cite{Georgiou2015, Hasegawa2017}. However, pain could also be interpreted as a proxy for establishing a high level of consciousness, which has been correlated with better outcomes in the ICU~\cite{Bastos1993}. Other top ranking phenotypes included tachycardia, atrial arrhythmia, constitutional symptom, nausea and vomiting, abnormal pupillary function, abnormal pattern of respiration, abnormal prothrombin time, confusion, increased inflammatory response, abnormal respiratory system morphology, diarrhea, and dyspnea; that together cover most of the body systems (i.e., heart, lungs, GI tract, central nervous system, coagulation, infection, kidneys) which are typically assessed through clinically validated scores, including early warning scores (e.g., NEWS) and mortality scores (e.g., Charlson Comorbodity Index\cite{Charlson1987}, Elixhauser Index\cite{Elixhauser1998}, APACHE, SAPS).

Unexpectedly and interestingly, the top ranking feature for mortality prediction is whether the patient experiences pain or not. We observe also that the second top ranking feature is \textit{Constitutional symptom (HP:0025142)}. Noting this is actually the resulting phenotype after aggregating all of its children, this phenotype should be interpreted not as a textual mention in the patient's EHR of the broad term, but rather as a mention of any of its children (most notably generalized pain). Consequently, the second top feature again highlights the importance of pain.

Although not decisive, there is some initial evidence corroborating the fact that pain management improves outcomes in the ICU~\cite{Georgiou2015}. However, pain could also be interpreted as a proxy for establishing a high level of consciousness, which has been correlated with better outcomes in the ICU~\cite{Bastos1993}.

The other top ranking phenotypes, such as atrial arrhythmia, and nausea and vomiting, cover most of the body systems (i.e., heart, lungs, GI tract, central nervous system, coagulation, infection, kidneys) which are typically assessed through clinically validated scores e.g., APACHE, SAPS.

Our study also showed that though phenotypic features are not as important for decompensation as for in-hospital mortality (only 3 out of the top 20 features for this task were phenotypic ones), they are still useful because they provide a better estimation of the predicted risk. Given that this task concerns predicting mortality within the next 24 hours, bedside measurements become more informative thanks to their temporal correlation (also shown in Figure \ref{fig:task_per_bucket}). Nevertheless, bedside measurements can be ambiguous or provide an incomplete picture of the patient's status without the data found in clinical notes. 
% Thus, these notes can give a baseline on which the prediction can be based. 
For example, for one patient \textit{Neoplasm of the respiratory system (HP:0100606)} was found to be the top feature, and although this phenotype is persistent, it increases appropriately the risk of decompensation, giving overall a better estimation. An illustration of this patient is shown by Figure \ref{fig:decomp_heatmap}.
% This situation is illustrated by Figure \ref{fig:decomp_heatmap}, where the top ranking feature is a phenotype (i.e., HP:0100606, Neoplasm of the respiratory system). Even though this phenotype is marked as persistent and thus remains present throughout the ICU stay, it increases appropriately the risk of decompensation, giving overall a better estimation. 
% This fact is further confirmed by the calibration curve for decompensation (Figure \ref{fig:calibration_decomp_lstm}), evidencing that the inclusion of phenotypic information leads to consistently more calibrated estimates, and by Figure \ref{fig:bucket_decomp}, where it is observed that predictions for physiological decompensation benefit the most from phenotypes when the patient stays in the ICU for more than one week.

Similarly, the top features for long length-of-stay (more than 1 week) are presented in Figure \ref{fig:top_features_los} where we notice 10 of 20 top features are phenotypes.

\subsection*{Calibration}

\begin{figure}[!htbp]
     \centering
     \begin{subfigure}[b]{0.45\textwidth}
         \centering
         \includegraphics[width=\textwidth]{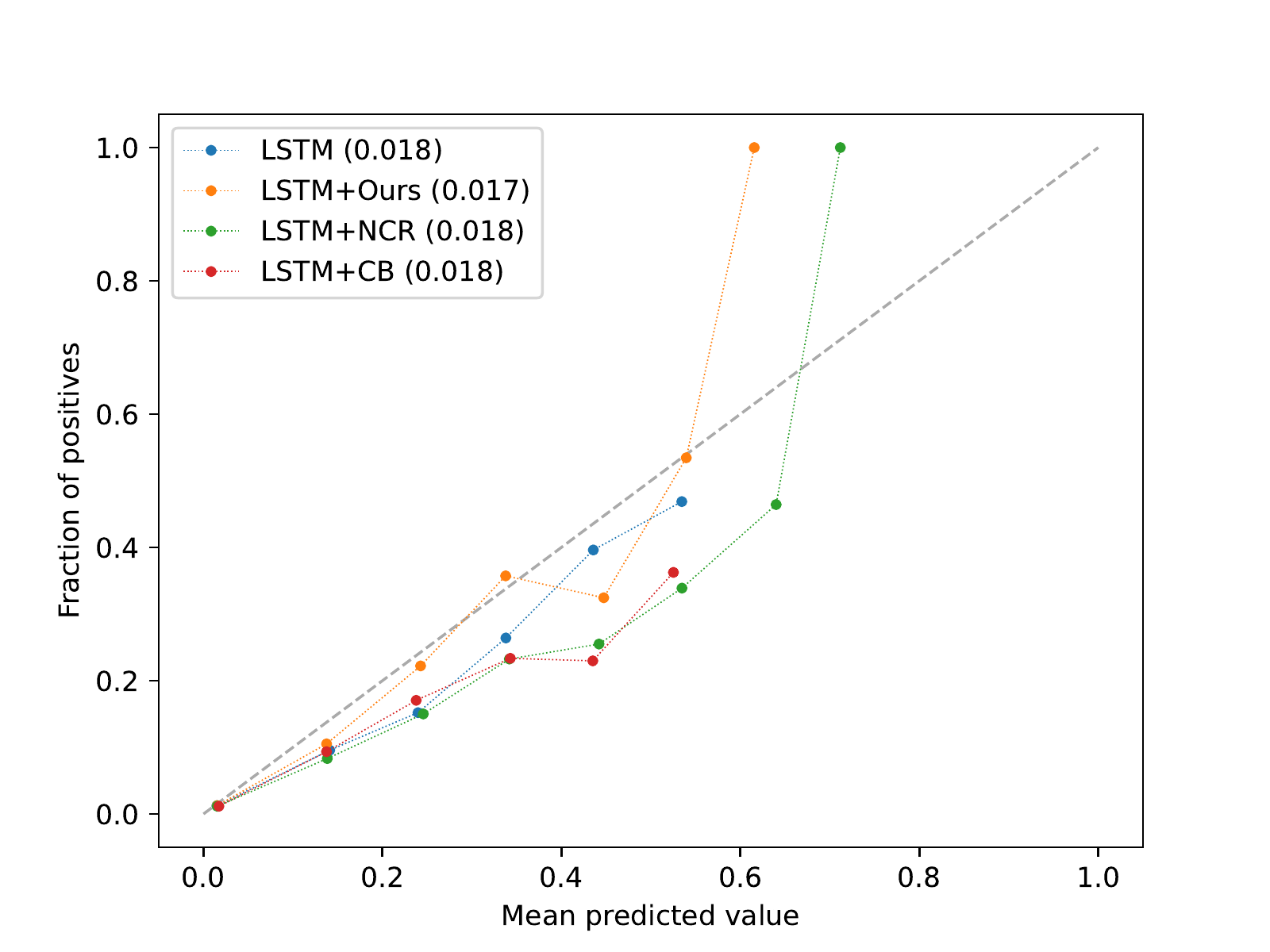}
         \caption{Physiological Decompensation}
         \label{fig:calibration_decomp_lstm}
     \end{subfigure}
     \hfill
     \begin{subfigure}[b]{0.45\textwidth}
         \centering
         \includegraphics[width=\textwidth]{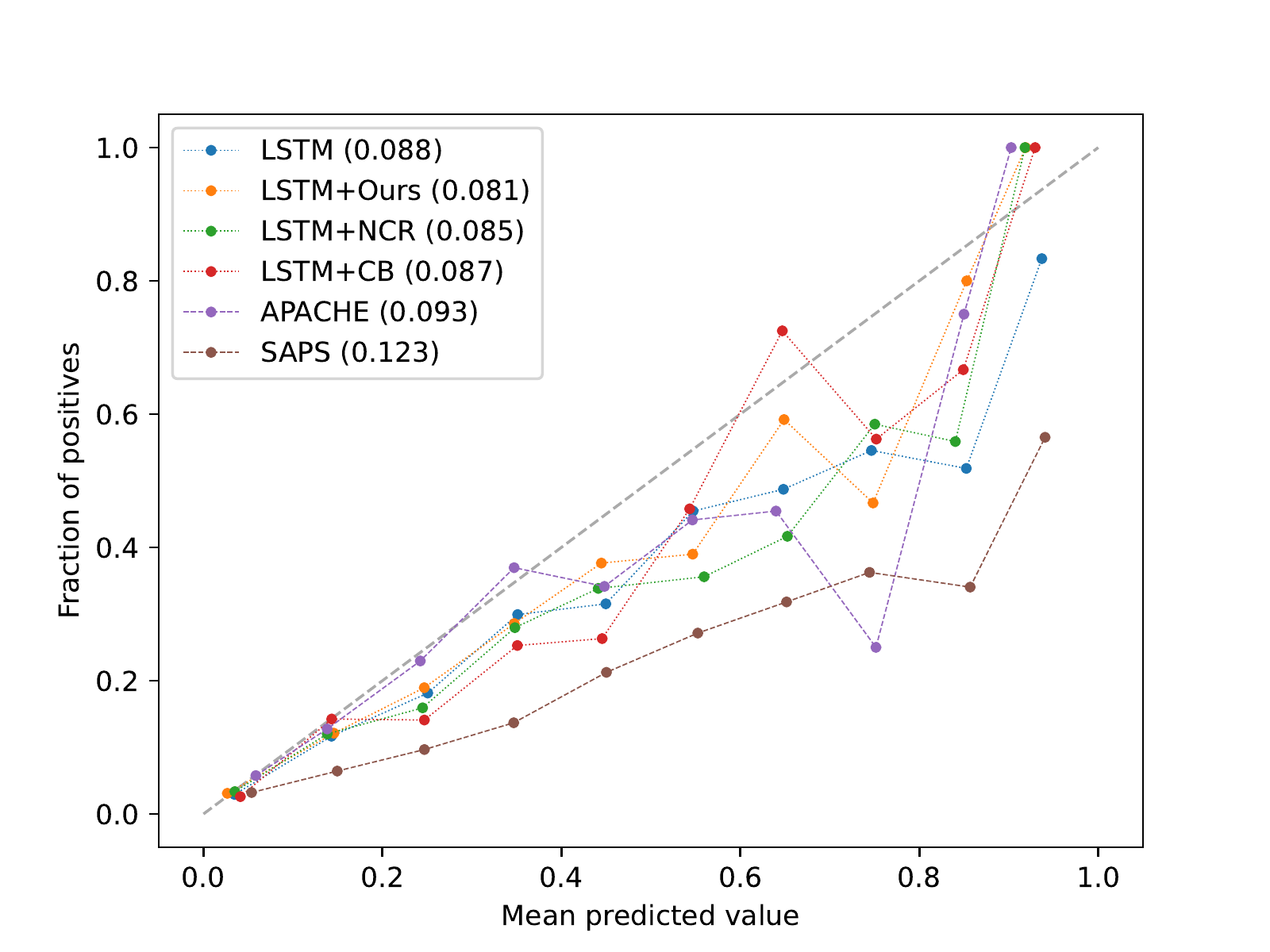}
         \caption{In-hospital Mortality}
         \label{fig:calibration_mortality_lstm}
     \end{subfigure}
    %  \hfill
        \caption{Calibration curves with LSTM for (a) physiological decompensation and (b) in-hospital mortality. Calibration curves are presented with its Brier score (the lower the better). Note that overall inclusion of phenotypic features from unstructured data helps with calibration. LSTM in legend refers to using structured features only. Ours, NCR, CB: phenotypic features from our phenotyping model, NCR and ClinicalBERT, respectively.}
        \label{fig:calibration_lstm}
\end{figure}

Calibration of machine learning models compares the distribution of the probability predicted by models with the distribution of probabilities observed in real data (e.g. real patients). To measure model calibration, we use the Brier score\cite{Brier1950} (the lower the better). Our investigation of the respective calibration curves (see Figure \ref{fig:calibration_lstm} and Figure \ref{fig:calibration_rf}) shows that phenotypes from unstructured notes improve model calibration across setups, especially for physiological decompensation and in-hospital mortality, which means the distribution predicted by models is closer to real distribution of patients. Besides, LSTM overall also produces better calibration than RF.

% We also notice physiological decompensation models are overall better calibrated, which is expected given that physiological decompensation has a near-future prediction scope (i.e., next 24 hours) compared to in-hospital mortality having no fixed time window. 

% APACHE and SAPS (clinically validated mortality scores) are worse or only comparable to our weakest settings (i.e., only structured data). 

% Results for RF are presented in Figure \ref{fig:calibration_rf}. Overall, calibration with RFs for in-hospital mortality is also comparable. However, RFs for physiological decompensation present significantly worse calibrations.

\subsection*{Prognosis analysis}

\begin{figure}[!htbp]
    \centering
    \includegraphics[width=\textwidth]{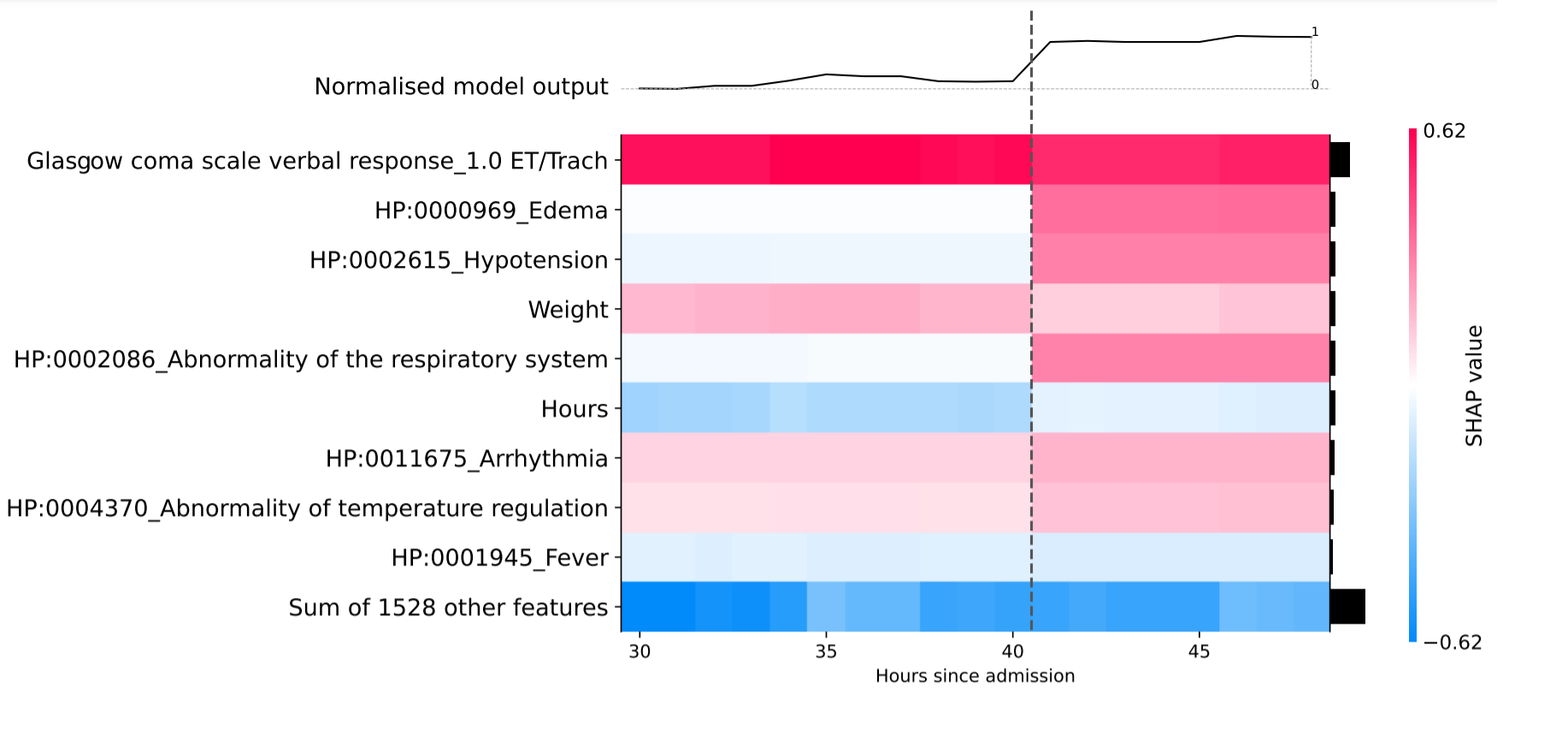}
    \medskip
    \dotfill\par
    \vspace{15pt}
    \centering
    \includegraphics[width=\textwidth]{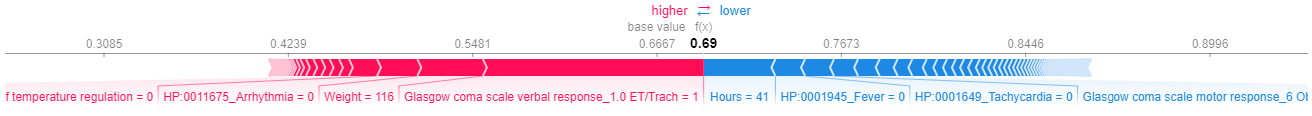}
    \vspace{0.5cm}
    \includegraphics[width=\textwidth]{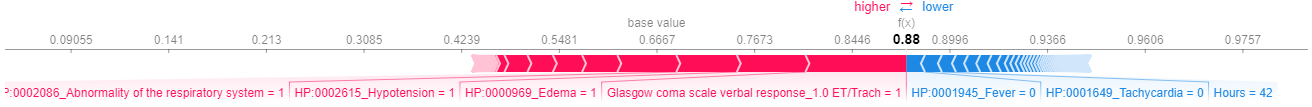}
    \caption{Illustrative case for an ICU length of stay of more than 14 days. \textbf{Top plot:} time course of the normalised predicted probability for a stay of more than 14 days, and feature heatmap for a representative segment of the ICU stay. Each row of the heatmap represents one of the top features. At each time step, a feature can contribute positively (red) or negatively (blue) for predicting a stay of 14 days or more. Black horizontal bars at the right of each row represent the importance of the features. Note that a new clinical note that is available at the 42nd hour (vertical dashed line) leads to an increase confidence of longer stay due to new features. \textbf{Bottom plot:} Inspection of the contributing features at the 41st and 42nd hours. Each force plot illustrates features contributing positively (red) and negatively (blue) to the prediction of a longer stay. Probability of long stay increases from 69\% to 88\% when the clinical note provides critical information.}
    \label{fig:los_before_after}
\end{figure}

Beyond producing clinically relevant explanations at the cohort level, with the help of SHAP values we can shed light onto a patient's journey and discover retrospectively when the patient was the most vulnerable and why. For example, the fragment of a patient's LOS forecast in Figure \ref{fig:los_before_after} illustrates an estimated probability, after 41 hour from admission, of a LOS longer than 14 days being of 69\%, mainly because the patient scored 1 in the Glasgow Coma Scale Verbal Response. One hour after, when a clinical note becomes available, worrisome phenotypes appear (including edema, hypotension and abnormality of the respiratory system). Consequently, the estimated probability increases to 88\%. 
% While hypotension is a concept the classifier can learn from the blood pressure measurements, edema and abnormality of the respiratory system are not reflected on typical acute measurements, all of which indicate the usefulness of phenotypic information on these tasks.

\subsection*{Cohort study}

We assess performance of our approach to the cohorts of the patients with different diseases especially underrepresented diseases to understand its robustness and generalisability. The test set is split into four disease-specific cohorts for patients with cardiovascular diseases, diabetes, cancer, and depression, and then the accuracies of the best LSTM models (using structured features and phenotypic features) are reported individually for each cohort on each ICU task. We notice the patient number of cardiovascular diseases or diabetes is at least twice that of cancer and around five times that of depression.
% We assess the approach's performance when tested against specific patients' cohorts, such as \textit{Diabetes, Cancer, Cardiovascular Diseases, and Depression}.
% We split our test set in these disease-specific cohorts and compute the scores using our best classifiers based on LSTM trained with both structured and unstructured data obtained by our phenotypes extraction methodology. Results are shown in Table \ref{table:results_generalizability} for all the tasks. They show that our approachs are robust to different cohorts of patients, including under-represented ones.

For in-hospital mortality and physiological decompensation, we observe comparable accuracies across the four cohorts. We report the range of AUC-ROC between 0.780 and 0.826 for in-hospital mortality and between 0.792 and 0.820 for physiological decompensation for the four cohorts. In contrast, for LOS, we observe lower Kappa 0.321 and 0.330 for small cohorts cancer and depression, respectively, as opposed to 0.413 and 0.424 for larger cohorts with cardiovascular diseases and diabetes. We hypothesise the nature of diseases has strong implication on in-hospital mortality and physiological decompensation while LOS can be influenced by more factors which require larger data samples to model their interactions. The full results are available in Table \ref{table:results_generalizability}.

% On the physiological decompensation task, we find all the cohorts show similar performance though all the scores are lower than the complete test set. So, our model appears to generalise well irrespective of the cohorts. Similarly on the in-hospital mortality task, the model generalises to all the cohorts with Cardiovascular Diseases and Depression getting lowest scores among all the cohorts. 

% Contrary, on the LOS task, it generalises to largest cohorts, i.e., Cardiovascular Diseases and Diabetes, well but the smallest cohorts, i.e., Cancer and Depression, show poor generalisability. It seems that for the physiological decompensation and in-hospital mortality tasks, nature of diseases affect the performance more than the sample size of the test set. For LOS task, smallest cohorts lag the performance which can be related to their sample size.

\section*{Limitations}

We have investigated only one data source, MIMIC-III, and our observations are to be confirmed with other data sources.

The analysis on phenotype importance is produced on the Random Forest, whose accuracy is superior than the baselines but not as good as LSTM. This is limited by the poor computation efficiency of SHAP values for LSTM and the explanations from neural network based models are to be investigated in future studies.

Moreover, the phenotypes annotated as transient are made present only until a new clinical note appears in the timeline. This has the inconvenience that phenotypes might be prematurely considered as not present because the next available clinical note did not mention them. Even though the LSTM classifier is able to learn temporal correlations on its own, a more elaborated feature modelling could prove useful.

\section*{Acknowledgements}

We would like to thank Dr. Rick Sax, Dr. Garima Gupta and Dr. Matt Wiener for their feedback throughout this research. We would also like to thank Dr. Garima Gupta, Dr. Deepa (M.R.S.H) and Dr. Ashok (M.S.) for helping us create gold-standard phenotype annotation data.

\section*{Author contributions}

JZ, LB, AT and JI conceived the experiments. LB and AT conducted the experiments. JZ, LB, AT, and JI analysed the results. JI, VB and YG reviewed the research and manuscript. All authors approved the manuscript.

\section*{Additional information}

\textbf{Competing interests}: This study is under collaboration with Imperial College London, Queen Mary University of London, Hong Kong Baptist University and Pangaea Data Limited.

\bibliography{references}

\appendix
\renewcommand{\thefigure}{A\arabic{figure}}
\renewcommand{\thetable}{A\arabic{table}}
\setcounter{figure}{0}
\setcounter{table}{0}
\newpage
\section*{Appendix}

\subsection*{Shapley Values}
\label{subsec:shap}

Shapley values come from game theory and are used to estimate the impact of a feature on a system's output. Feature impact is defined as the variation in the output of the model when the feature is observed versus when it is unknown. 

Shapley values belong to a category of methods denominated additive. In particular, the additivity is formulated as

\begin{equation*}
\begin{split}
    f(x) & = \phi_0(f,x) + \sum_{i=1}^{M}\phi_i(f,x)
\end{split}
\end{equation*}

where $f(x)$ is the prediction made by the model, $x$ are the features fed to the model, $M$ is the number of features, $\phi_i$ is the Shapley value of the i-th feature, and $\phi_0 = E[f(x)]$ is the expected value of the model over the training dataset. Also, this assumption ensures the values correctly reflect the difference between the expected model output and the output for a particular prediction.

The Shapley value of a feature is computed via

\begin{equation}
\begin{split}
 \phi_i(f,x) & = \sum_{S\subseteq S_{all}/\{i\}}\frac{|S|! (M-|S|-1)!}{M!}[f_x(S \cup \{i\}) - f_x(S)] \\
& = \sum_{S\subseteq S_{all}/\{i\}}\frac{1}{(M \text{ choose } |S|)(M-|S|)}[f_x(S \cup \{i\}) - f_x(S)]
\end{split}
\label{eq:shap_value}
\end{equation}

where $S$ is a subset of all $M$ input features, and $f_x(S) = E[f(x) | x_s]$ with $x_s$ in a subset of the input features with only those belonging to $S$ present.

In this study we used the SHAP library \cite{Lundberg2017original} and its optimisation for tree-based classifiers \cite{Lundberg2020tree} to compute the Shapley values. 

\newpage

\subsection{Data distribution by splits}

\begin{table}[htbp]
\centering
\begin{tabular}{|c|c|c|c|c|c|c|l}
\cline{1-7}
                                       & \multicolumn{6}{c|}{Counts}                                                                                                                                                                      & \multicolumn{1}{c}{} \\ \cline{1-7}
\multirow{2}{*}{Task}                  & \multirow{2}{*}{Split} & \multirow{2}{*}{Patients} & \multirow{2}{*}{ICU Episodes} & \multirow{2}{*}{Timesteps} & \multicolumn{2}{c|}{Labels}                                                      &                      \\ \cline{6-7}
                                       &                        &                           &                             &                            & Positive                                & Negative                               &                      \\ \cline{1-7}
\multirow{5}{*}{Physiological Decompensation}& CV-1                   & 5125                      & 6215                        & 528425                     & 10283                                   & 518142                                 &                      \\ \cline{2-7}
                                       & CV-2                   & 5129                      & 6134                        & 507892                     & 10821                                   & 497071                                 &                      \\ \cline{2-7}
                                       & CV-3                   & 5141                      & 6264                        & 511289                     & 10426                                   & 500863                                 &                      \\ \cline{2-7}
                                       & CV-4                   & 5102                      & 6297                        & 527853                     & 11020                                   & 516833                                 &                      \\ \cline{2-7}
                                       & Test                   & 3683                      & 4463                        & 367533                     & 6931                                    & 360602                                 &                      \\ \cline{1-7}
\multirow{5}{*}{In-Hospital Mortality} & CV-1                   & 2929                      & 3382                        & 162063                     & 441                                     & 2941                                   &                      \\ \cline{2-7}
                                       & CV-2                   & 2917                      & 3331                        & 159566                     & 466                                     & 2865                                   &                      \\ \cline{2-7}
                                       & CV-3                   & 2888                      & 3356                        & 160732                     & 439                                     & 2917                                   &                      \\ \cline{2-7}
                                       & CV-4                   & 2936                      & 3410                        & 163284                     & 477                                     & 2933                                   &                      \\ \cline{2-7}
                                       & Test                   & 2119                      & 2453                        & 117500                     & 283                                     & 2170                                   &                      \\ \cline{1-7}
\multirow{5}{*}{Length of Stay}        & CV-1                   & 5151                      & 6245                        & 532403                     & \multicolumn{2}{c|}{\multirow{5}{*}{Refer to Table \ref{table:los_classes}}} &                      \\ \cline{2-5}
                                       & CV-2                   & 5145                      & 6154                        & 510227                     & \multicolumn{2}{c|}{}                                                            &                      \\ \cline{2-5}
                                       & CV-3                   & 5160                      & 6286                        & 514147                     & \multicolumn{2}{c|}{}                                                            &                      \\ \cline{2-5}
                                       & CV-4                   & 5117                      & 6314                        & 530331                     & \multicolumn{2}{c|}{}                                                            &                      \\ \cline{2-5}
                                       & Test                   & 3698                      & 4483                        & 369350                     & \multicolumn{2}{c|}{}                                                            &                      \\ \cline{1-7}
\end{tabular}
\caption{Data distribution by splits. For the physiological decompensation and length of stay tasks, timesteps are taken as samples as the predictions are made every hourly timesteps, while for the in-hospital mortality task, ICU episodes are taken as samples as the predictions are made at a fixed timestep. Here, \textit{CV} refers to the training \textit{Cross-Validation} Folds.}
\label{table:data_distribution}
\end{table}

\subsection{Class distribution for length of stay}

\begin{table}[htbp]
\centering
\begin{tabular}{|c|c|c|c|c|c|c|}
\hline
Class   Label & Class Description (Days) & CV-1   & CV-2   & CV-3   & CV-4   & Test   \\ \hline
0             & \textless 1              & 131913 & 129634 & 131693 & 133186 & 95439  \\ \hline
1             & 1 - 2                    & 85311  & 83558  & 84065  & 85818  & 61372  \\ \hline
2             & 2 - 3                    & 56353  & 54074  & 54007  & 54780  & 38858  \\ \hline
3             & 3 - 4                    & 39416  & 37605  & 38106  & 38054  & 27142  \\ \hline
4             & 4 - 5                    & 29384  & 27982  & 28760  & 28573  & 20171  \\ \hline
5             & 5 - 6                    & 22830  & 22384  & 22360  & 22626  & 15878  \\ \hline
6             & 6 - 7                    & 18816  & 18612  & 18626  & 18582  & 12940  \\ \hline
7             & 7 - 8                    & 15925  & 15583  & 15697  & 15863  & 10953  \\ \hline
8             & 8 - 14                   & 62655  & 58512  & 59905  & 60611  & 40856  \\ \hline
9             & \textgreater 14          & 69800  & 62283  & 60928  & 72238  & 45741  \\ \hline
Total         &                          & 532403 & 510227 & 514147 & 530331 & 369350 \\ \hline
\end{tabular}
\caption{Class distribution for Length of Stay}
\label{table:los_classes}
\end{table}

\subsection*{Algorithm hyperparameters}

\begin{table}[H]
\centering
\begin{tabular}{|c|p{12cm}|}
\hline
Classifier                & Hyperparameters                                                                                                                   \\ \hline
Random Forest             & num of estimators=300, criterion="gini", max depth=None, min samples split=2, min samples leaf=1                                 \\ \hline
LSTM                      & epochs=30, hidden size=128, batch size=8, num of layers=1, patience=10, dropout rate=0, learning rate=1e-4, weight decay=0.0 \\ \hline
\end{tabular}
\caption{Hyperparameters for classifiers}
\label{table:hyperarameters}
\end{table}

\newpage
\subsection*{Significance tests}

\begin{table}[H]
\begin{subtable}[c]{1\textwidth}
\centering
\scalebox{1.0}{
\begin{tabular}{|c|c|c|c|c|c|}
\hline
\multirow{2}{*}{\begin{tabular}[c]{@{}c@{}}ML \\ Classification \\ Model\end{tabular}} & \multirow{2}{*}{\begin{tabular}[c]{@{}c@{}}Base \\ Model\end{tabular}} & \multicolumn{4}{c|}{Secondary Models}                                                                                                                             \\ \cline{3-6} 
                                                                                       &                                                                        & S   & \begin{tabular}[c]{@{}c@{}}S + \\ NCR\end{tabular} & \begin{tabular}[c]{@{}c@{}}S + \\ CB\end{tabular} & \begin{tabular}[c]{@{}c@{}}S + \\ Ours\end{tabular}\\ \hline
\multirow{4}{*}{\begin{tabular}[c]{@{}c@{}}Random\\ Forest\end{tabular}}               & S                                                                      & -   & 1                                                  & 13.25                                             & 0                                               \\ \cline{2-6} 
                                                                                       & S + NCR                                                                & 99  & -                                                  & 94.82                                             & 26.13                                              \\ \cline{2-6} 
                                                                                       & S + CB                                                                 & 86.75 & 5.18                                             & -                                                 & 1.07                                               \\ \cline{2-6} 
                                                                                       & S + Ours                                                               & 100 & 73.87                                            & 98.93                                             & -                                                  \\ \hline
% \multirow{4}{*}{\begin{tabular}[c]{@{}c@{}}Gradient\\ Boosting\end{tabular}}           & S                                                                      & -   & 0.6                                                & 1.78                                              & 2.26                                               \\ \cline{2-6} 
%                                                                                       & S + NCR                                                                & 99.4& -                                                  & 88.39                                             & 84.43                                              \\ \cline{2-6} 
%                                                                                       & S + CB                                                                 & 98.22 & 11.61                                            & -                                                 & 44.07                                              \\ \cline{2-6} 
%                                                                                       & S + Ours                                                               & 97.74 & 15.57                                            & 55.93                                             & -                                                  \\ \hline
\multirow{4}{*}{LSTM}                                                                  & S                                                                      & -   & 0                                                  & 0                                                 & 0                                                  \\ \cline{2-6} 
                                                                                       & S + NCR                                                                & 100 & -                                                  & 100                                               & 0                                                  \\ \cline{2-6} 
                                                                                       & S + CB                                                                 & 100 & 0                                                  & -                                                 & 0                                                  \\ \cline{2-6} 
                                                                                       & S + Ours                                                               & 100 & 100                                                & 100                                               & -                                                  \\ \hline
\end{tabular}
}
\subcaption{In-Hospital Mortality}
\end{subtable}
\bigskip
\begin{subtable}[c]{1\textwidth}
\centering
\scalebox{1.0}{
\begin{tabular}{|c|c|c|c|c|c|}
\hline
\multirow{2}{*}{\begin{tabular}[c]{@{}c@{}}ML \\ Classification \\ Model\end{tabular}} & \multirow{2}{*}{\begin{tabular}[c]{@{}c@{}}Base \\ Model\end{tabular}} & \multicolumn{4}{c|}{Secondary Models}                                                                                                                               \\ \cline{3-6} 
                                                                                       &                                                                        & S    & \begin{tabular}[c]{@{}c@{}}S + \\ NCR\end{tabular} & \begin{tabular}[c]{@{}c@{}}S + \\ CB\end{tabular} & \begin{tabular}[c]{@{}c@{}}S + \\ Ours\end{tabular} \\ \hline
\multirow{4}{*}{\begin{tabular}[c]{@{}c@{}}Random \\ Forest\end{tabular}}              & S                                                                      & -    & 81.4                                               & 69                                                & 0                                                \\ \cline{2-6} 
                                                                                       & S + NCR                                                                & 18.6 & -                                                  & 32.4                                              & 0                                                 \\ \cline{2-6} 
                                                                                       & S + CB                                                                 & 31   & 67.6                                               & -                                                 & 0                                                 \\ \cline{2-6} 
                                                                                       & S + Ours                                                               & 100 & 100                                               & 100                                              & -                                                   \\ \hline
% \multirow{4}{*}{\begin{tabular}[c]{@{}c@{}}Gradient \\ Boosting\end{tabular}}          & S                                                                      & -    & 9.5                                                & 0                                                 & 0                                                   \\ \cline{2-6} 
%                                                                                       & S + NCR                                                                & 90.5 & -                                                  & 6.1                                               & 0                                                   \\ \cline{2-6} 
%                                                                                       & S + CB                                                                 & 100  & 93.9                                               & -                                                 & 0.1                                                 \\ \cline{2-6} 
%                                                                                       & S + Ours                                                               & 100  & 100                                                & 99.9                                              & -                                                   \\ \hline
\multirow{4}{*}{LSTM}                                                                  & S                                                                      & -    & 0                                                  & 0                                                 & 0                                                   \\ \cline{2-6} 
                                                                                       & S + NCR                                                                & 100  & -                                                  & 73                                                & 0                                                   \\ \cline{2-6} 
                                                                                       & S + CB                                                                 & 100  & 27                                                 & -                                                 & 0                                                   \\ \cline{2-6} 
                                                                                       & S + Ours                                                               & 100  & 100                                                & 100                                               & -                                                   \\ \hline
\end{tabular}
}
\subcaption{Physiological Decompensation}
\end{subtable}
\bigskip
\begin{subtable}[c]{1\textwidth}
\centering
\scalebox{1.0}{
\begin{tabular}{|c|c|c|c|c|c|}
\hline
\multirow{2}{*}{\begin{tabular}[c]{@{}c@{}}ML \\ Classification \\ Model\end{tabular}} & \multirow{2}{*}{\begin{tabular}[c]{@{}c@{}}Base \\ Model\end{tabular}} & \multicolumn{4}{c|}{Secondary Models}                                                                                                                              \\ \cline{3-6} 
                                                                                       &                                                                        & S    & \begin{tabular}[c]{@{}c@{}}S + \\ NCR\end{tabular} & \begin{tabular}[c]{@{}c@{}}S + \\ CB\end{tabular} & \begin{tabular}[c]{@{}c@{}}S + \\ Ours\end{tabular} \\ \hline
\multirow{4}{*}{\begin{tabular}[c]{@{}c@{}}Random\\ Forest\end{tabular}}               & S                                                                      & -    & 22.1                                               & 100                                               & 0                                                  \\ \cline{2-6} 
                                                                                       & S + NCR                                                                & 77.9 & -                                                  & 100                                               & 0                                                  \\ \cline{2-6} 
                                                                                       & S + CB                                                                 & 0    & 0                                                  & -                                                 & 0                                                  \\ \cline{2-6} 
                                                                                       & S + Ours                                                               & 100  & 100                                                & 100                                               & -                                                  \\ \hline
% \multirow{4}{*}{\begin{tabular}[c]{@{}c@{}}Gradient\\ Boosting\end{tabular}}           & S                                                                      & -    & 100                                                & 100                                               & 0                                                  \\ \cline{2-6} 
%                                                                                       & S + NCR                                                                & 0    & -                                                  & 0                                                 & 0                                                  \\ \cline{2-6} 
%                                                                                       & S + CB                                                                 & 0    & 100                                                & -                                                 & 0                                                  \\ \cline{2-6} 
%                                                                                       & S + Ours                                                               & 100  & 100                                                & 100                                               & -                                                  \\ \hline
\multirow{4}{*}{LSTM}                                                                  & S                                                                      & -    & 0                                                  & 0                                                 & 0                                                  \\ \cline{2-6} 
                                                                                       & S + NCR                                                                & 100  & -                                                  & 100                                               & 0                                                  \\ \cline{2-6} 
                                                                                       & S + CB                                                                 & 100  & 0                                                  & -                                                 & 0                                                  \\ \cline{2-6} 
                                                                                       & S + Ours                                                               & 100  & 100                                                & 100                                               & -                                                  \\ \hline
\end{tabular}}
\subcaption{Length of Stay}
\end{subtable}
\bigskip

\caption{Statistical Significance Matrix with Bootstrap Resampling. All the scores are percentages of samples where the base model performs better than the secondary model. Each sample is built by resampling the original test set and then scoring the base/secondary model on it. For example, the last row in (a) shows the base model (LSTM with S + Ours) is better than the secondary models (LSTM with S or S + NCR or S + CB) on 100\% samples (i.e. with statistical significance). Here, S refers to Structured, NCR to Neural Concept Recognizer\cite{arbabi2019ncr}, CB to ClinicalBERT, and Ours to our phenotyping model. }
\label{table:results_stat_test}
\end{table}

\newpage
\subsection*{4-Fold cross validation results}

\begin{table}[!htbp]
\begin{subtable}[c]{1\textwidth}
\centering
\begin{tabular}{|c|c|cccc|}
\hline
\multirow{3}{*}{Classification Model} & \multirow{2}{*}{Features Design} & \multicolumn{4}{c|}{4-Fold Cross Validation Aggregate}                                                         \\ \cline{3-6} 
                                      &                                  & \multicolumn{2}{c|}{AUC-ROC}                                     & \multicolumn{2}{c|}{AUC-PR}                 \\ \cline{2-6} 
                                      &                                  & \multicolumn{1}{c|}{Mean}           & \multicolumn{1}{c|}{SD}    & \multicolumn{1}{c|}{Mean}           & SD    \\ \hline
SAPS-II                               & -                                & \multicolumn{1}{c|}{0.754}          & \multicolumn{1}{c|}{0.006} & \multicolumn{1}{c|}{0.322}          & 0.031 \\ \hline
APACHE-III                            & -                                & \multicolumn{1}{c|}{0.732}          & \multicolumn{1}{c|}{0.008} & \multicolumn{1}{c|}{0.326}          & 0.018 \\ \hline
\multirow{4}{*}{Random   Forest}      & S                                & \multicolumn{1}{c|}{0.810}          & \multicolumn{1}{c|}{0.008} & \multicolumn{1}{c|}{0.418}          & 0.018 \\ \cline{2-6} 
                                      & S + NCR                          & \multicolumn{1}{c|}{0.819}          & \multicolumn{1}{c|}{0.014} & \multicolumn{1}{c|}{0.472}          & 0.013 \\ \cline{2-6} 
                                      & S + CB                           & \multicolumn{1}{c|}{0.804}          & \multicolumn{1}{c|}{0.012} & \multicolumn{1}{c|}{0.423}          & 0.005 \\ \cline{2-6} 
                                      & S + Ours                         & \multicolumn{1}{c|}{\textbf{0.834}} & \multicolumn{1}{c|}{0.008} & \multicolumn{1}{c|}{\textbf{0.477}} & 0.016 \\ \hline
\multirow{5}{*}{LSTM}                 & S                                & \multicolumn{1}{c|}{-}              & \multicolumn{1}{c|}{-}     & \multicolumn{1}{c|}{-}              & -     \\ \cline{2-6} 
                                      & S                                & \multicolumn{1}{c|}{0.829}          & \multicolumn{1}{c|}{0.007} & \multicolumn{1}{c|}{0.441}          & 0.016 \\ \cline{2-6} 
                                      & S + NCR                          & \multicolumn{1}{c|}{0.836}          & \multicolumn{1}{c|}{0.011} & \multicolumn{1}{c|}{0.478}          & 0.008 \\ \cline{2-6} 
                                      & S + CB                           & \multicolumn{1}{c|}{0.829}          & \multicolumn{1}{c|}{0.007} & \multicolumn{1}{c|}{0.459}          & 0.007 \\ \cline{2-6} 
                                      & S + Ours                         & \multicolumn{1}{c|}{\textbf{0.845}} & \multicolumn{1}{c|}{0.004} & \multicolumn{1}{c|}{\textbf{0.496}} & 0.014 \\ \hline
\end{tabular}
\caption{In-hospital mortality}
\label{table:mortality_cv}
\end{subtable}

\begin{subtable}[c]{1\textwidth}
\centering
\begin{tabular}{|c|c|cccc|}
\hline
\multirow{3}{*}{Classification Model} & \multirow{3}{*}{Features Design} & \multicolumn{4}{c|}{4-Fold Cross Validation Aggregate}                                                         \\ \cline{3-6} 
                                      &                                  & \multicolumn{2}{c|}{AUC-ROC}                                     & \multicolumn{2}{c|}{AUC-PR}                 \\ \cline{3-6} 
                                      &                                  & \multicolumn{1}{c|}{Mean}           & \multicolumn{1}{c|}{SD}    & \multicolumn{1}{c|}{Mean}           & SD    \\ \hline
\multirow{4}{*}{Random   Forest}      & S                                & \multicolumn{1}{c|}{0.815}          & \multicolumn{1}{c|}{0.003} & \multicolumn{1}{c|}{0.127}          & 0.009 \\ \cline{2-6} 
                                      & S + NCR                          & \multicolumn{1}{c|}{0.820}          & \multicolumn{1}{c|}{0.003} & \multicolumn{1}{c|}{0.125}          & 0.007 \\ \cline{2-6} 
                                      & S + CB                           & \multicolumn{1}{c|}{0.818}          & \multicolumn{1}{c|}{0.004} & \multicolumn{1}{c|}{0.123}          & 0.008 \\ \cline{2-6} 
                                      & S + Ours                         & \multicolumn{1}{c|}{\textbf{0.844}} & \multicolumn{1}{c|}{0.004} & \multicolumn{1}{c|}{\textbf{0.165}} & 0.013 \\ \hline
\multirow{5}{*}{LSTM}                 & S                                & \multicolumn{1}{c|}{-}              & \multicolumn{1}{c|}{-}     & \multicolumn{1}{c|}{-}              & -     \\ \cline{2-6} 
                                      & S                                & \multicolumn{1}{c|}{0.819}          & \multicolumn{1}{c|}{0.003} & \multicolumn{1}{c|}{0.136}          & 0.016 \\ \cline{2-6} 
                                      & S + NCR                          & \multicolumn{1}{c|}{0.820}          & \multicolumn{1}{c|}{0.003} & \multicolumn{1}{c|}{0.134}          & 0.013 \\ \cline{2-6} 
                                      & S + CB                           & \multicolumn{1}{c|}{0.821}          & \multicolumn{1}{c|}{0.006} & \multicolumn{1}{c|}{0.128}          & 0.022 \\ \cline{2-6} 
                                      & S + Ours                         & \multicolumn{1}{c|}{\textbf{0.833}} & \multicolumn{1}{c|}{0.008} & \multicolumn{1}{c|}{\textbf{0.144}} & 0.023 \\ \hline
\end{tabular}
\caption{Physiological decompensation}
\label{table:decompensation_cv}
\end{subtable}

\begin{subtable}[c]{1\textwidth}
\centering
\begin{tabular}{|c|c|cccc|}
\hline
\multirow{3}{*}{Classification Model} & \multirow{3}{*}{Features Design} & \multicolumn{4}{c|}{4-Fold Cross Validation Aggregate}                                                            \\ \cline{3-6} 
                                      &                                  & \multicolumn{2}{c|}{Kappa}                                       & \multicolumn{2}{c|}{MAD}                       \\ \cline{3-6} 
                                      &                                  & \multicolumn{1}{c|}{Mean}           & \multicolumn{1}{c|}{SD}    & \multicolumn{1}{c|}{Mean}             & SD     \\ \hline
\multirow{4}{*}{Random Forest}        & S                                & \multicolumn{1}{c|}{0.381}          & \multicolumn{1}{c|}{0.005} & \multicolumn{1}{c|}{142.010}          & 4.665  \\ \cline{2-6} 
                                      & S + NCR                          & \multicolumn{1}{c|}{0.382}          & \multicolumn{1}{c|}{0.008} & \multicolumn{1}{c|}{148.003}          & 4.180  \\ \cline{2-6} 
                                      & S + CB                           & \multicolumn{1}{c|}{0.369}          & \multicolumn{1}{c|}{0.005} & \multicolumn{1}{c|}{149.221}          & 3.789  \\ \cline{2-6} 
                                      & S + Ours                         & \multicolumn{1}{c|}{\textbf{0.405}} & \multicolumn{1}{c|}{0.006} & \multicolumn{1}{c|}{\textbf{116.940}} & 5.674  \\ \hline
\multirow{5}{*}{LSTM}                 & S                                & \multicolumn{1}{c|}{-}              & \multicolumn{1}{c|}{-}     & \multicolumn{1}{c|}{-}                & -      \\ \cline{2-6} 
                                      & S                                & \multicolumn{1}{c|}{0.375}          & \multicolumn{1}{c|}{0.003} & \multicolumn{1}{c|}{134.373}          & 17.293 \\ \cline{2-6} 
                                      & S + NCR                          & \multicolumn{1}{c|}{0.393}          & \multicolumn{1}{c|}{0.013} & \multicolumn{1}{c|}{127.165}          & 17.484 \\ \cline{2-6} 
                                      & S + CB                           & \multicolumn{1}{c|}{0.374}          & \multicolumn{1}{c|}{0.015} & \multicolumn{1}{c|}{127.678}          & 8.608  \\ \cline{2-6} 
                                      & S + Ours                         & \multicolumn{1}{c|}{\textbf{0.416}} & \multicolumn{1}{c|}{0.012} & \multicolumn{1}{c|}{\textbf{116.198}} & 6.904  \\ \hline
\end{tabular}
\caption{Length of Stay}
\label{table:los_cv}
\end{subtable}

\caption{Results for (a) In-Hospital Mortality, (b) Physiological Decompensation, and (c) Length of Stay on the training set. The best score for each classifier is highlighted in bold. Here, S refers to Structured, NCR to Neural Concept Recognizer\cite{arbabi2019ncr}, CB to ClinicalBERT, and Ours to our phenotyping model.}
\label{table:results_cross_validation}
\end{table}

\newpage
\subsection{Ablation study on phenotype persistency}
\label{appendix:ablation_study}

\begin{table}[htbp]
\begin{subtable}[c]{1\textwidth}
\centering
\begin{tabular}{|c|c|c|c|c|c|c|c|}
\hline
\multirow{3}{*}{Model} & \multirow{3}{*}{\begin{tabular}[c]{@{}c@{}}Phenotypic \\ propagation\end{tabular}} & \multicolumn{4}{c|}{4-fold Cross Validation Aggregate}        & \multicolumn{2}{c|}{Test Set}   \\ \cline{3-8} 
                       &                                         & \multicolumn{2}{c|}{AUC-ROC}      & \multicolumn{2}{c|}{AUC-PR}       & \multirow{2}{*}{AUC-ROC}         & \multirow{2}{*}{AUC-PR}          \\ \cline{3-6}
                       &                                         & Mean            & SD              & Mean            & SD              &                                  &                                  \\ \hline
\multirow{2}{*}{RF}    & without                                 & 0.807          & 0.008          & 0.413          & 0.021          & 0.799 (0.772, 0.824)          & 0.351 (0.297, 0.407)          \\ \cline{2-8} 
                       & with                                    & \textbf{0.834} & 0.008 & \textbf{0.477} & 0.016 & \textbf{0.845 (0.826, 0.873)} & \textbf{0.462 (0.404, 0.524)} \\ \hline
\multirow{2}{*}{LSTM}  & without                                 & 0.833          & 0.014          & 0.457          & 0.024          & 0.831 (0.807, 0.853)          & 0.421 (0.361, 0.483)          \\ \cline{2-8} 
                       & with                                    & \textbf{0.844} & 0.004 & \textbf{0.495} & 0.013 & \textbf{0.845 (0.823, 0.868)} & \textbf{0.464 (0.405, 0.523)} \\ \hline
\end{tabular}
\caption{In-hospital Mortality}
\label{table:ablation_mortality}
\end{subtable}

\begin{subtable}[c]{1\textwidth}
\centering
\begin{tabular}{|c|c|c|c|c|c|c|c|}
\hline
\multirow{3}{*}{Model} & \multirow{3}{*}{\begin{tabular}[c]{@{}c@{}}Phenotypic \\ propagation\end{tabular}} & \multicolumn{4}{c|}{4-fold Cross Validation Aggregate}         & \multicolumn{2}{c|}{Test Set}    \\ \cline{3-8} 
                       &                                         & \multicolumn{2}{c|}{AUC-ROC}      & \multicolumn{2}{c|}{AUC-PR}        & \multirow{2}{*}{AUC-ROC}         & \multirow{2}{*}{AUC-PR}          \\ \cline{3-6}
                       &                                         & Mean            & SD              & Mean            & SD               &                                  &                                  \\ \hline
\multirow{2}{*}{RF}    & without                                 & 0.812          & 0.002          & 0.125         & 0.007           & 0.820 (0.815, 0.825)          & 0.127 (0.120, 0.135) \\ \cline{2-8} 
                       & with                                    & \textbf{0.844} & 0.004 & \textbf{0.165} & 0.013  & \textbf{0.845 (0.840, 0.850)} & \textbf{0.180 (0.171, 0.190)}          \\ \hline
\multirow{2}{*}{LSTM}  & without                                 & 0.827          & 0.007          & \textbf{0.146} & 0.017 & \textbf{0.841 (0.842, 0.851)} & \textbf{0.149 (0.141, 0.156)} \\ \cline{2-8} 
                       & with                                    & \textbf{0.833} & 0.007 & 0.144          & 0.022           & 0.839 (0.834, 0.844)          & 0.145 (0.138, 0.153)          \\ \hline
\end{tabular}
\caption{Physiological Decompensation}
\label{table:ablation_decomp}
\end{subtable}

\begin{subtable}[c]{1\textwidth}
\centering
\begin{tabular}{|c|c|c|c|c|c|c|c|}
\hline
\multirow{3}{*}{Model} & \multirow{3}{*}{\begin{tabular}[c]{@{}c@{}}Phenotypic \\ propagation\end{tabular}} & \multicolumn{4}{c|}{4-fold Cross Validation Aggregate}          & \multicolumn{2}{c|}{Test Set}           \\ \cline{3-8} 
                       &                                         & \multicolumn{2}{c|}{Kappa}      & \multicolumn{2}{c|}{MAD}            & \multirow{2}{*}{Kappa}         & \multirow{2}{*}{MAD}                    \\ \cline{3-6}
                       &                                         & Mean            & SD              & Mean              & SD              &                                  &                                         \\ \hline
\multirow{2}{*}{RF}    & without                                 & 0.376          & 0.005          & 139.8          & 5.5 & 0.386 (0.380, 0.384)          & 135.0 (134.5, 135.6)  \\ \cline{2-8} 
                       & with                                    & \textbf{0.405} & 0.006 & \textbf{116.9} & 5.6          & \textbf{0.420 (0.418, 0.422)} & \textbf{110.3 (109.3, 111.3)}           \\ \hline
\multirow{2}{*}{LSTM}  & without                                 & \textbf{0.427} & 0.007 & 118.3 & 4.2 & \textbf{0.441 (0.439, 0.440)} & \textbf{111.4 (110.9, 111.9)}  \\ \cline{2-8} 
                       & with                                    & 0.416          & 0.012          & \textbf{116.2}          & 6.9          & 0.430 (0.427, 0.432)          & 116.7 (116.2, 117.3)           \\ \hline
\end{tabular}
\caption{Length of Stay}
\label{table:ablation_los}
\end{subtable}
\caption{Results of ablation study on our phenotyping model to assess the importance of phenotypic modelling. Models without phenotypic propagation encounter high sparsity of phenotypes as data is only available at the timestep the clinical note is written. Models with phenotypic propagation observe phenotypes throughout all timesteps. The best score for each classifier is highlighted in bold.}
\label{table:ablation}
\end{table}

\newpage
\subsection*{Feature importance for Length-of-Stay}

\begin{figure}[H]
    \centering
    \includegraphics[width=\textwidth]{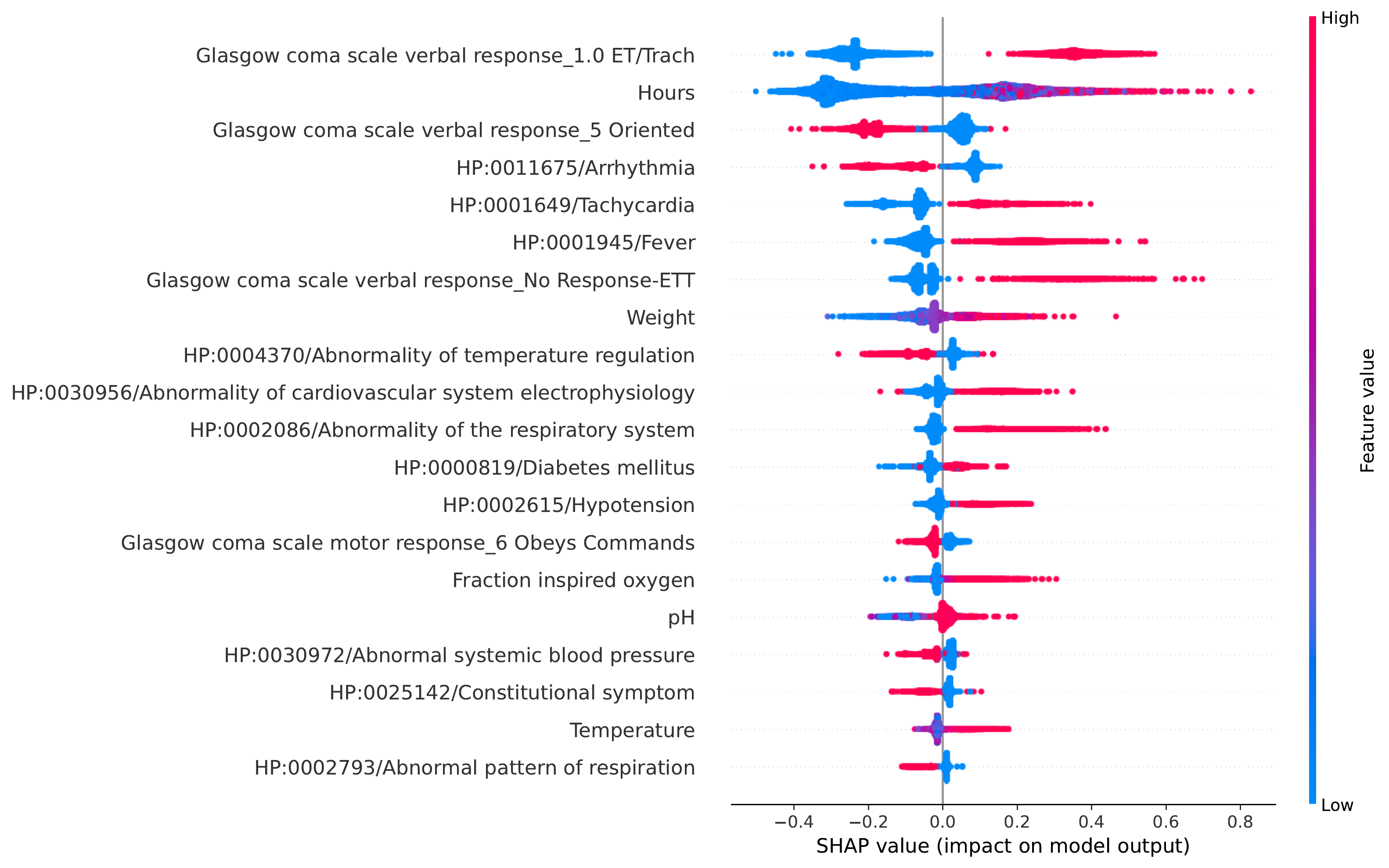}
    \caption{Top features for length-of-stay predicting stays of more than 1 week.}
    \label{fig:top_features_los}
\end{figure}

\subsection*{Calibration curves}
\label{appendix:calibration_curves}

% \begin{figure}[htbp]
%      \centering
%      \begin{subfigure}[b]{0.45\textwidth}
%          \centering
%          \includegraphics[width=\textwidth]{figures/calibration_decomp_all_GB.pdf}
%          \caption{Physiological Decompensation}
%          \label{fig:calibration_decomp_gb}
%      \end{subfigure}
%      \hfill
%      \begin{subfigure}[b]{0.45\textwidth}
%          \centering
%          \includegraphics[width=\textwidth]{figures/calibration_mortality_all_GB.pdf}
%          \caption{In-hospital Mortality}
%          \label{fig:calibration_mortality_gb}
%      \end{subfigure}
%     %  \hfill
%         \caption{Calibration curves with Gradient Boosting for (a) physiological decompensation and (b) in-hospital mortality.}
%         \label{fig:calibration_gb}
% \end{figure}

\begin{figure}[htbp]
     \centering
     \begin{subfigure}[b]{0.45\textwidth}
         \centering
         \includegraphics[width=\textwidth]{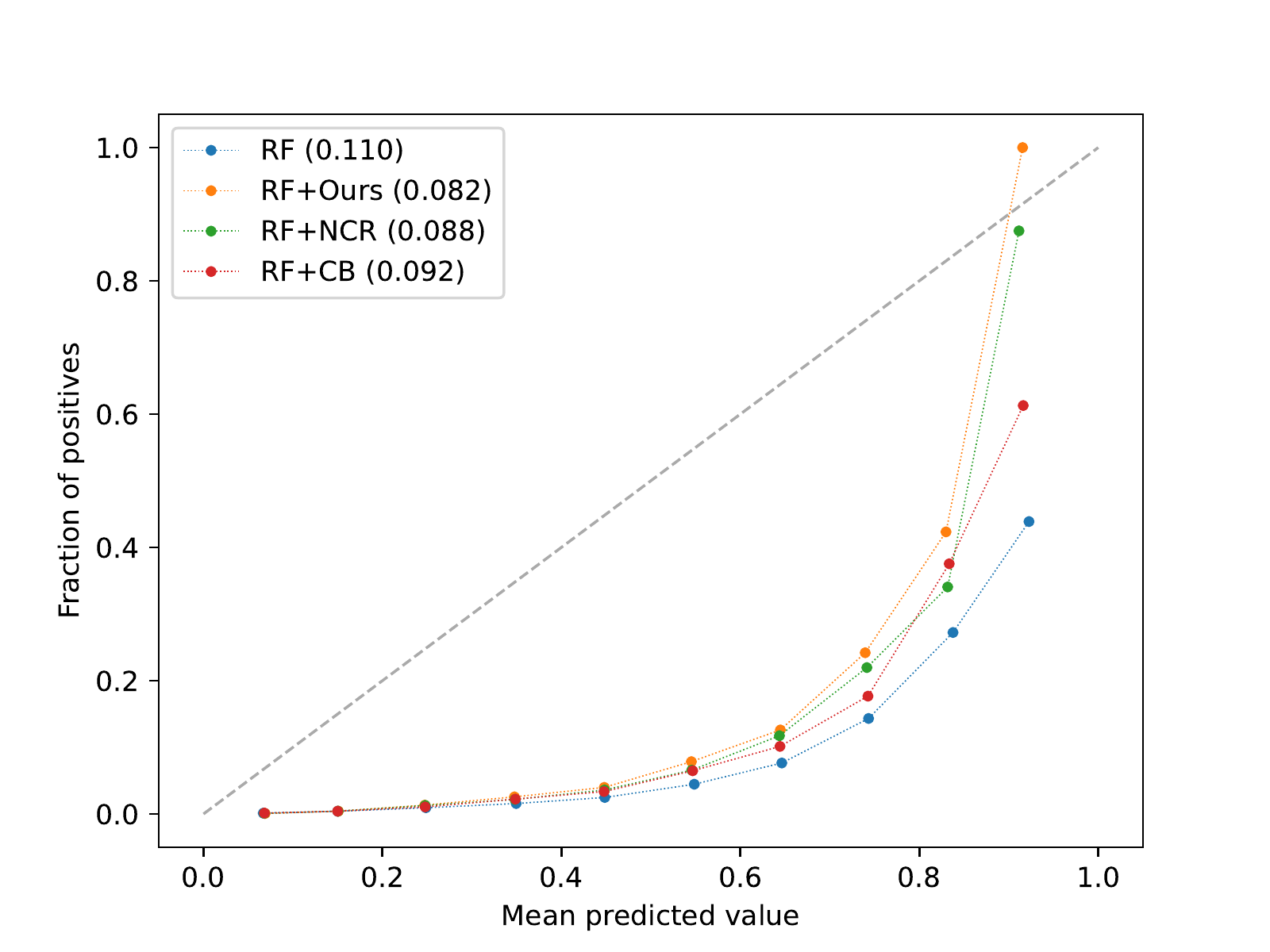}
         \caption{Physiological Decompensation}
         \label{fig:calibration_decomp_rf}
     \end{subfigure}
     \hfill
     \begin{subfigure}[b]{0.45\textwidth}
         \centering
         \includegraphics[width=\textwidth]{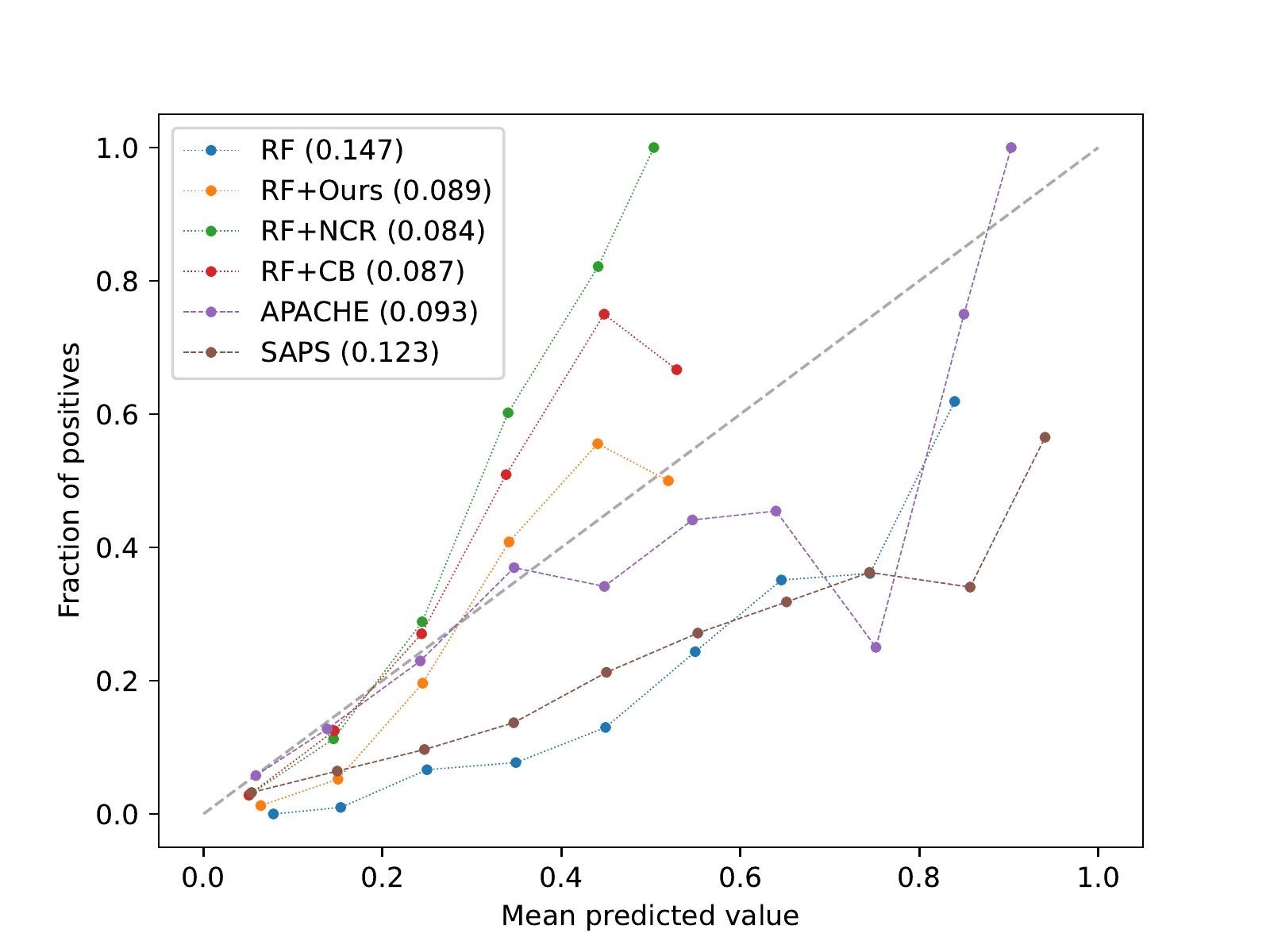}
         \caption{In-hospital Mortality}
         \label{fig:calibration_mortality_rf}
     \end{subfigure}
    %  \hfill
        \caption{Calibration curves with Random Forest for (a) physiological decompensation and (b) in-hospital mortality. RF in legend refers to using structured features only. Ours, NCR, CB: phenotypic features from our phenotyping model, NCR and ClinicalBERT, respectively.}
        \label{fig:calibration_rf}
\end{figure}

\newpage
\subsection{Cohort study}

\begin{table}[!htbp]

\begin{subtable}[c]{1\textwidth}
\centering
\scalebox{0.8}{
\begin{tabular}{|c|c|c|c|}
\hline
Cohort             & No. of Patients & No. of ICU Episodes & AUC-ROC \\ \hline
All                & 2119            & 2453              & 0.845  \\ \hline
Cardiovascular Diseases             & 681             & 789               & 0.780  \\ \hline
Diabetes           & 563             & 682               & 0.826  \\ \hline
Cancer             & 277             & 304               & 0.822  \\ \hline
Depression         & 119             & 122               & 0.783   \\ \hline
% Multiple Sclerosis & 19              & 21                & 0.5     \\ \hline
\end{tabular}}
\subcaption{In-hospital Mortality.}
\end{subtable}

\bigskip
\begin{subtable}[c]{1\textwidth}
\centering
\scalebox{0.8}{
\begin{tabular}{|c|c|c|c|}
\hline
Cohort             & No. of Patients & No. of ICU Episodes & AUC-ROC \\ \hline
All                & 3683            & 4463              & 0.839  \\ \hline
Cardiovascular Diseases             & 975             & 1197              & 0.792  \\ \hline
Diabetes           & 927             & 1191              & 0.808  \\ \hline
Cancer             & 489             & 565               & 0.806  \\ \hline
Depression         & 216             & 240               & 0.820  \\ \hline
% Multiple Sclerosis & 26              & 33                & 0.570  \\ \hline
\end{tabular}
}
\subcaption{Physiological Decompensation.}
\end{subtable}

\bigskip
\begin{subtable}[c]{1\textwidth}
\centering
\scalebox{0.80}{
\begin{tabular}{|c|c|c|c|}
\hline
Cohort             & No. of Patients & No. of ICU Episodes & Kappa \\ \hline
All                & 3698            & 4483              & 0.430      \\ \hline
Cardiovascular Diseases             & 980             & 1202              & 0.413      \\ \hline
Diabetes           & 930             & 1195              & 0.424      \\ \hline
Cancer             & 493             & 572               & 0.321      \\ \hline
Depression         & 216             & 241               & 0.330      \\ \hline
% Multiple Sclerosis & 26              & 34                & 0.538       \\ \hline
\end{tabular}
}
\subcaption{Length of Stay}
\end{subtable}

\caption{Analysing the generalisability and robustness of our approach on cohorts with different diseases. The accuracies of the best LSTM models which use features from both structured and unstructured data are reported individually on each cohort for each ICU task.}
\label{table:results_generalizability}
\end{table}

\newpage
\subsection{Forecasts per total length of stay}

\begin{figure}[!htbp]
     \centering
     \begin{subfigure}[b]{0.45\textwidth}
         \centering
         \includegraphics[width=\textwidth]{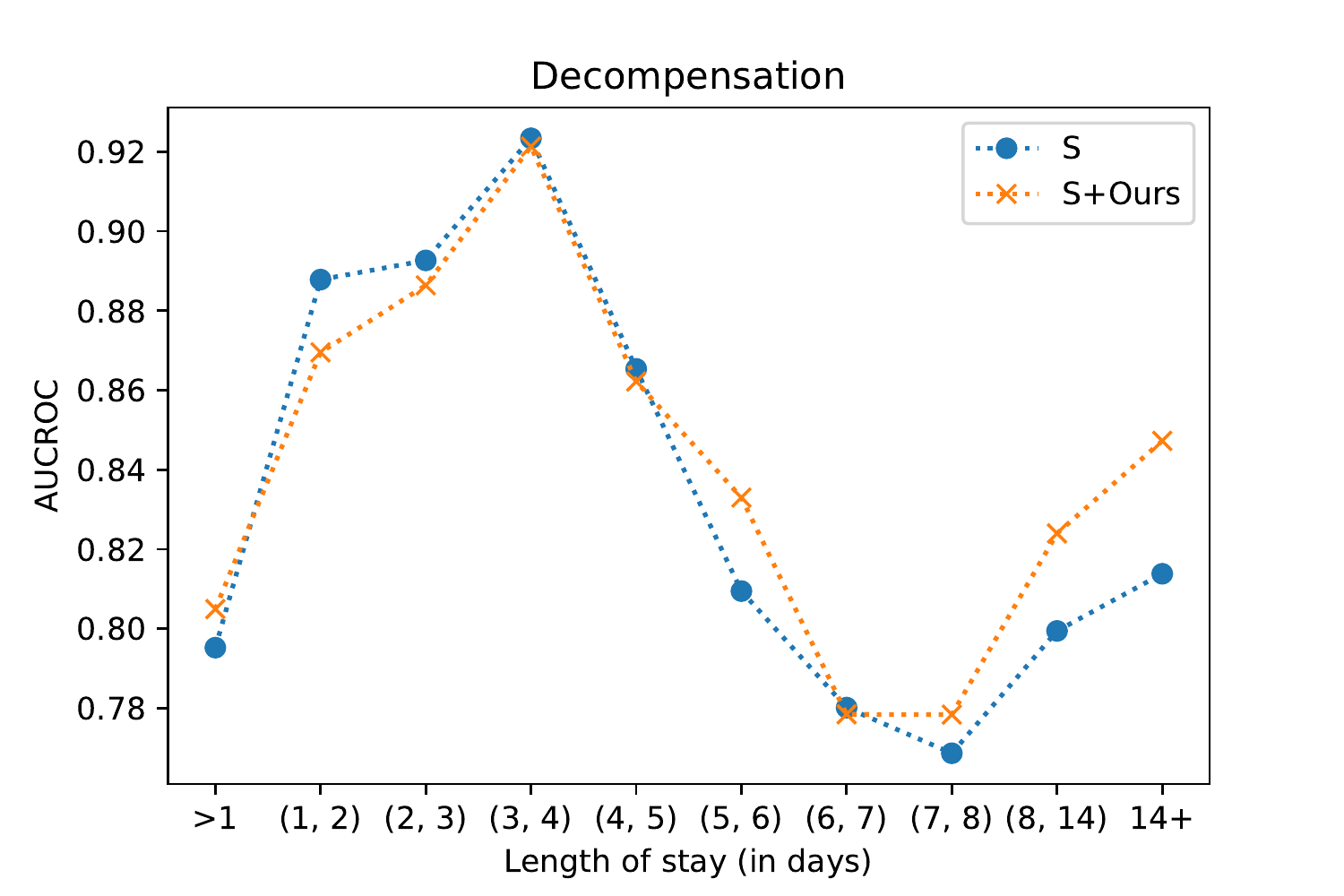}
         \caption{Physiological Decompensation}
         \label{fig:bucket_decomp}
     \end{subfigure}
     \hfill
     \begin{subfigure}[b]{0.45\textwidth}
         \centering
         \includegraphics[width=\textwidth]{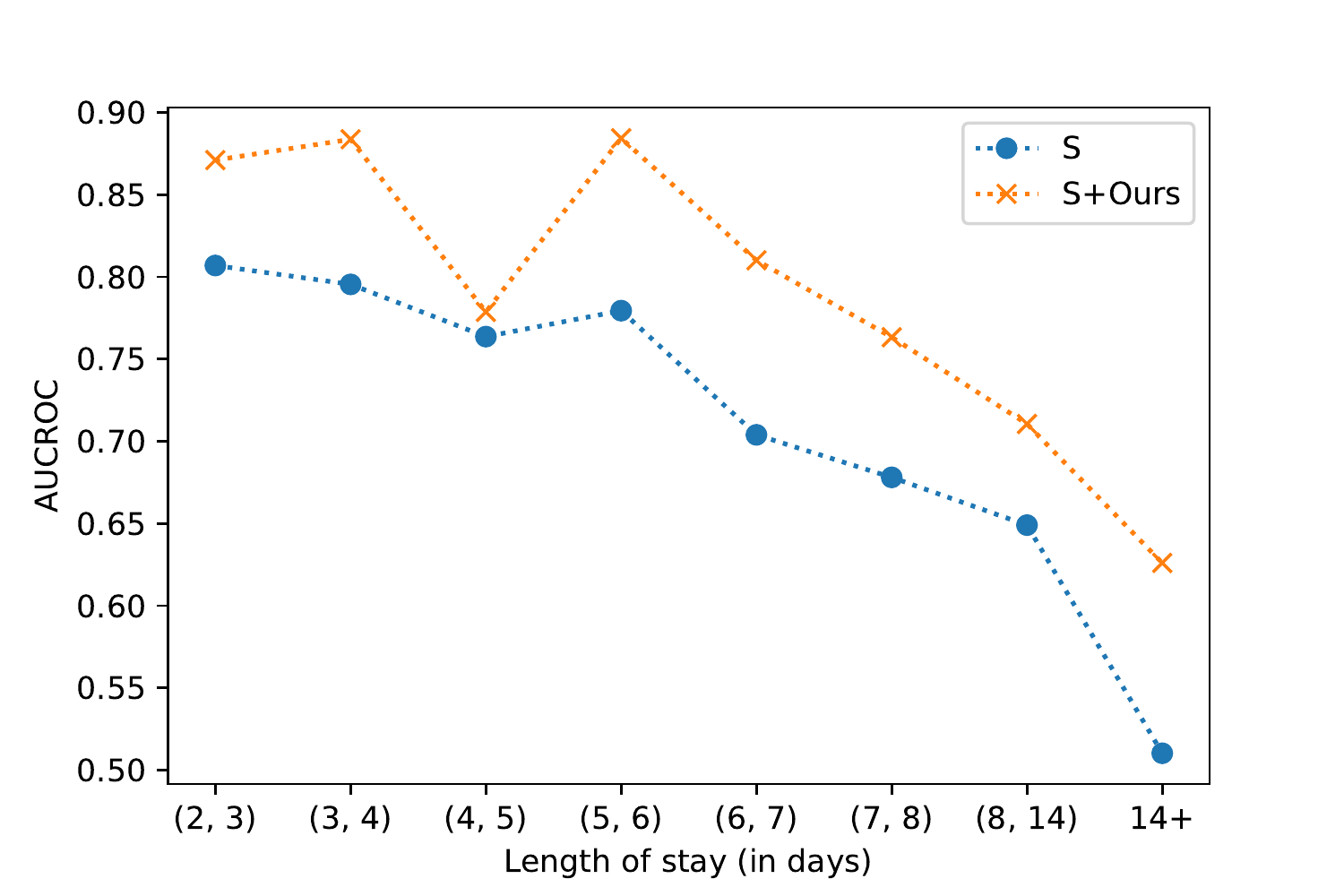}
         \caption{In-hospital Mortality}
         \label{fig:bucket_mortality}
     \end{subfigure}
    %  \hfill
        \caption{AUCROC for (a) physiological decompensation and (b) in-hospital mortality for LSTM for patients with different LOS values. While the in-hospital mortality task benefits consistently for any duration of the ICU stay, decompensation sees the best improvements when patients stay the longest. This behaviour is a natural consequence of the fact that while near future forecasts can rely strongly on bedside measurements, forecasting without a fixed endpoint in time is significantly more difficult. Nevertheless, patients who stayed for less than two weeks still saw a benefit when introducing phenotypic features, as they calibrate better the algorithm's prediction. Here, S represents structured features and Ours refers to phenotypes from our phenotyping model.}
        \label{fig:task_per_bucket}
\end{figure}

\subsection{Case study for physiological decompensation}

\begin{figure}[!htbp]
    \centering
    \includegraphics[width=\textwidth]{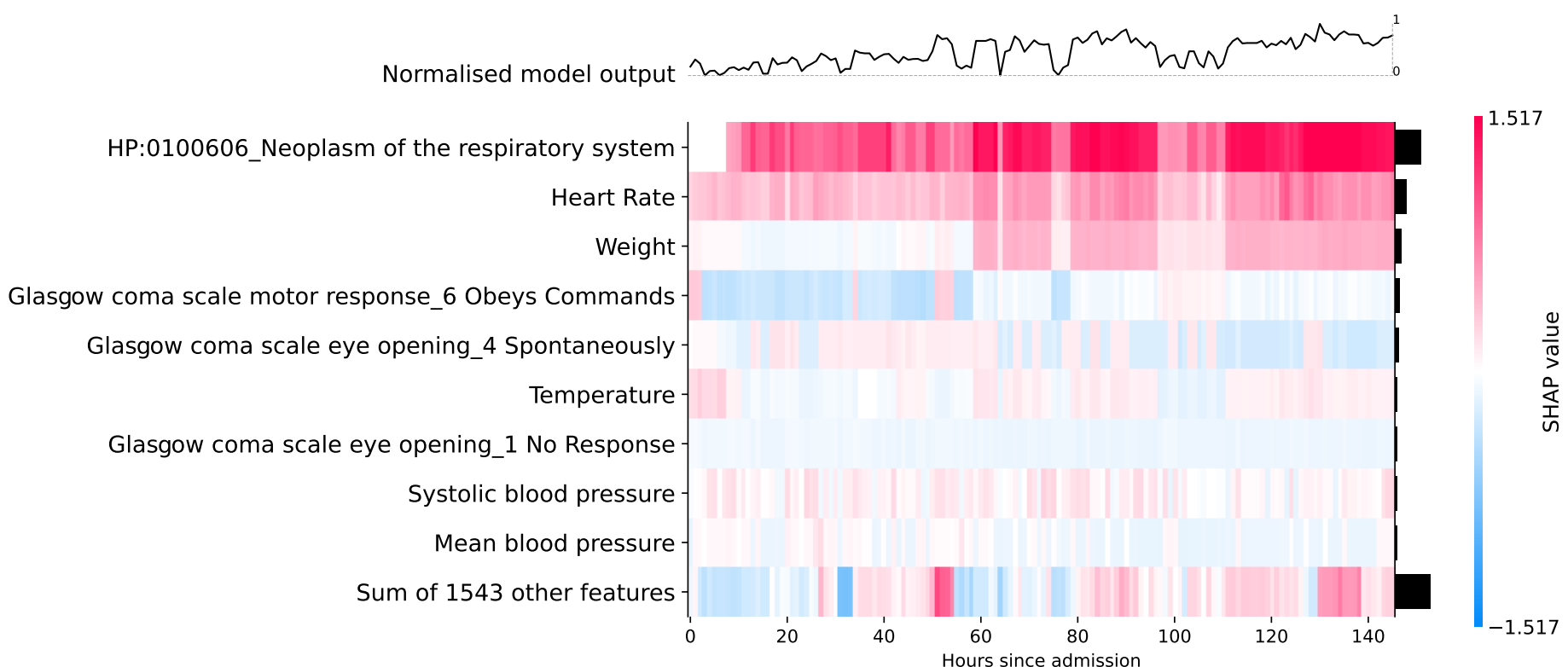}
    \caption{Time course of the physiological decompensation prediction for an illustrative patient in the test set. The top plot represents the time series of the prediction in probability (0 for no risk of decompensation, 1 for decompensation). The heatmap illustrates how the contribution of each feature (i.e., each row) varies across time for this subject. Features are sorted in decreasing order according to their importance for this patient, represented by the black horizontal bar at the right of each row. The colour of a row indicates how that feature contributes to the prediction at a moment in time, with red representing a positive contribution (i.e., that the patient will decompensate), and blue for a negative contribution. For this patient, although fluctuations in the prediction come from changes in structured data, taking into account the neoplasm of the respiratory system allows to better estimate the baseline risk of decompensation.}
    \label{fig:decomp_heatmap}
\end{figure}

\end{document}